\documentclass{article}

\usepackage[final]{neurips}

\usepackage[utf8]{inputenc}
\usepackage[T1]{fontenc}
\usepackage{hyperref}
\usepackage{url}
\usepackage{booktabs}
\usepackage{amsfonts}
\usepackage{nicefrac}
\usepackage{microtype}
\usepackage[dvipsnames]{xcolor}
\usepackage[many]{tcolorbox}
\usepackage{amssymb}
\usepackage{colortbl}
\usepackage{pifont}
\usepackage{multirow}
\usepackage{graphicx}
\usepackage{tabularx}
\usepackage{threeparttable}
\usepackage{wrapfig}
\usepackage{enumitem}
\usepackage{subcaption}
\usepackage{gradient-text}

\newcommand{\cmark}{\ding{51}}
\newcommand{\xmark}{\ding{55}}

\newcommand{\ie}{\textit{i.e.}}
\newcommand{\eg}{\textit{e.g.}}

\newcommand{\model}{UniPixel}
\newcommand{\modelgradient}{\gradientRGB{UniPixel}{236,47,75}{0,159,255}}

\newcommand{\email}[1]{\texttt{\href{mailto:#1}{#1}}}

\author{
Ye Liu$^{1,2}$, Zongyang Ma$^{2,3}$, Junfu Pu$^2$, Zhongang Qi$^4$, Yang Wu$^5$,\\\textbf{Ying Shan}$^2$\textbf{,} \textbf{Chang Wen Chen}$^1$\thanks{\textit{Corresponding author.}} \\
$^1$\,The Hong Kong Polytechnic University\; $^2$\,ARC Lab, Tencent PCG\; \\
$^3$\,Institute of Automation, Chinese Academy of Sciences\; $^4$\,vivo Mobile Communication Co. \\
$^5$\,MindWingman Technology (Shenzhen) Co., Ltd. \\
\email{coco.ye.liu@connect.polyu.hk} \\
\color{magenta}\url{https://polyu-chenlab.github.io/unipixel/}}

\title{\modelgradient: Unified Object Referring and \\ Segmentation for Pixel-Level Visual Reasoning}

\begin{document}

\setlength{\skip\footins}{15pt}

\maketitle

\begin{figure}[!h]
\vspace{-5mm}
\centering
\includegraphics[width=0.95\linewidth]{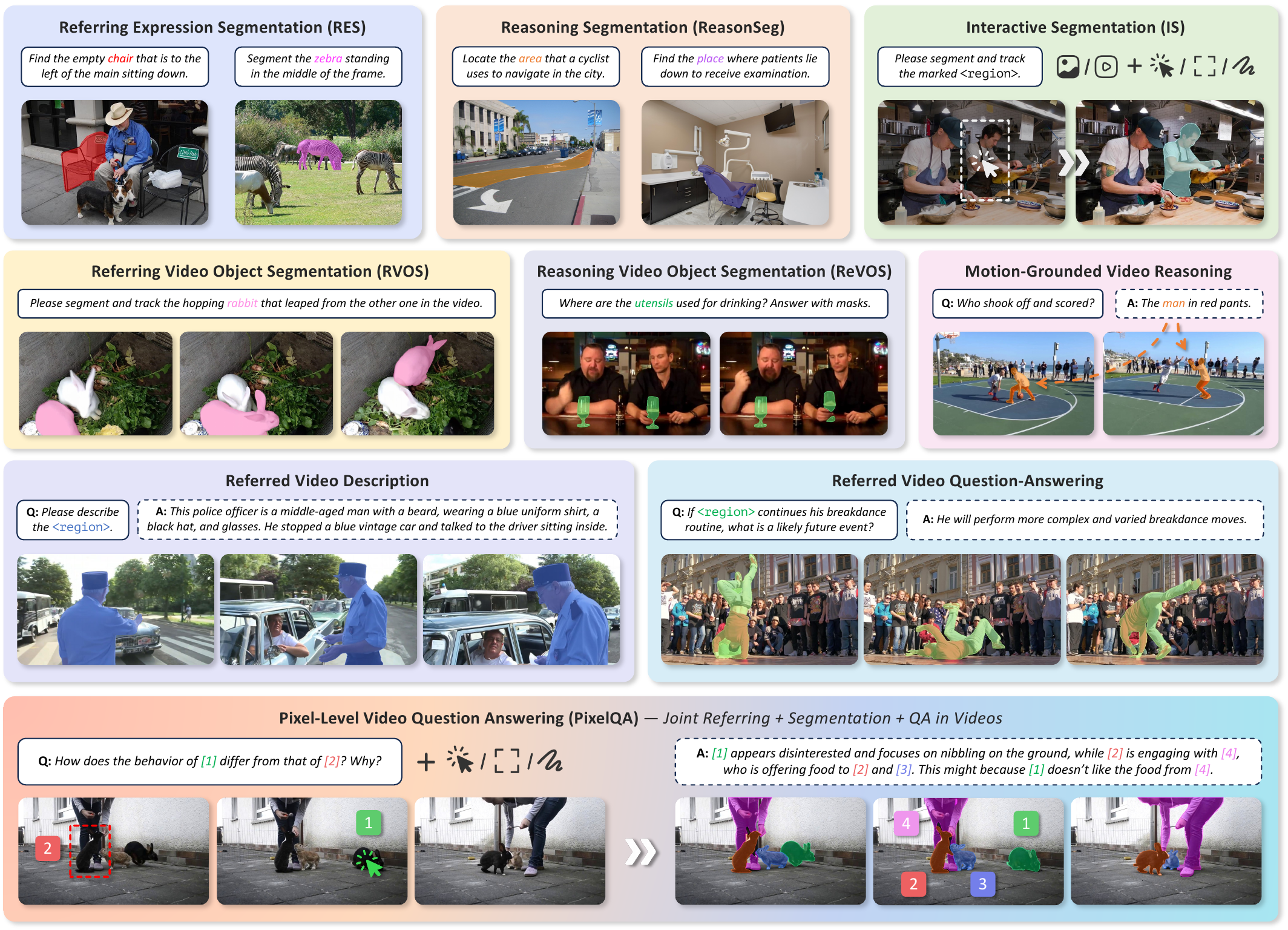}
\caption{\textbf{\model} flexibly supports a large variety of fine-grained image and video understanding tasks, including referring/reasoning/interactive segmentation, motion-grounded video reasoning, and referred video description \& question answering. It can also handle a novel \textbf{PixelQA} task that jointly requires object-centric referring, segmentation, and question answering in videos.}
\label{fig:task}
\end{figure}

\begin{abstract}
Recent advances in Large Multi-modal Models (LMMs) have demonstrated their remarkable success as general-purpose multi-modal assistants, with particular focuses on holistic image- and video-language understanding. Conversely, less attention has been given to scaling fine-grained pixel-level understanding capabilities, where the models are expected to realize pixel-level alignment between visual signals and language semantics. Some previous studies have applied LMMs to related tasks such as region-level captioning and referring expression segmentation. However, these models are limited to performing either referring or segmentation tasks independently and fail to integrate these fine-grained perception capabilities into visual reasoning. To bridge this gap, we propose \textbf{\model}, a large multi-modal model capable of flexibly comprehending visual prompt inputs and generating mask-grounded responses. Our model distinguishes itself by seamlessly integrating pixel-level perception with general visual understanding capabilities. Specifically, \model{} processes visual prompts and generates relevant masks on demand, and performs subsequent reasoning conditioning on these intermediate pointers during inference, thereby enabling fine-grained \textbf{pixel-level reasoning}. The effectiveness of our approach has been verified on 10 benchmarks across a diverse set of tasks, including pixel-level referring/segmentation and object-centric understanding in images/videos. A novel \textbf{PixelQA} task that jointly requires referring, segmentation, and question answering is also designed to verify the flexibility of our method.
\end{abstract}

\setlength{\skip\footins}{10pt}

\section{Introduction}

Large Multi-modal Models (LMMs) have been the de facto standard for developing general-purpose assistants. By effectively aligning multi-modalities with language, their significance has been demonstrated across various applications, including multimedia analysis \cite{gpt4o,gemini2.5,zhang2025aesthetics,zhang2025smaller,vlog,videomind}, autonomous driving (AD) \cite{llm4ad,omnidrive,chatscene,asyncdriver}, and Embodied AI \cite{matchaagent,palme,openvla,octopi}.

In the field of visual-language understanding, efforts have been dedicated to developing \textit{holistic understanding models}, where simple projection layers between visual encoders and LLMs are utilized to bridge vision and language modalities. Supported by large-scale alignment pre-training and visual instruction tuning, such a straightforward paradigm achieves strong performance in holistic understanding tasks such as captioning \cite{mscoco,activitynet,youcook2} and general question answering \cite{mvbench,videomme,egoschema,etbench}. However, these models exhibit two fundamental limitations in fine-grained scenarios. \textbf{First}, their interactions with users are limited to text format, lacking support for more intuitive forms of communication such as drawing points/boxes as references or grounding model responses with key regions represented by masks. \textbf{Second}, the internal reasoning process of these models predominantly operates at a coarse level, directly perceiving the entire content rather than reasoning over specific objects/regions, making them hard to understand fine-grained details. Some previous studies have explored the application of LMMs to related tasks such as region-level captioning \cite{gpt4roi,osprey,videorefer}, referring expression segmentation \cite{refcoco,refcocog,gres,mevis,urvos,davis2017}, and reasoning segmentation \cite{lisa,mmr,visa,videolisa,villa}. Nevertheless, their models are limited to performing either referring or segmentation tasks independently via rigidly defined input/output templates (\eg, ``It's \texttt{<SEG>}.'' in LISA \cite{lisa}), lacking the flexibility to comprehend user-referred concepts and generate mask-grounded responses simultaneously. More importantly, these methods cannot integrate such fine-grained perception capabilities with their original human-like \cite{singlepulsenc,singlepulse,eegfmri,shortsinglepulse,visualthalamus} multi-modal reasoning abilities, resulting in degraded performance on general visual understanding benchmarks \cite{activitynetqa,videoqa,tgifqa}.

In this work, we seek to bridge this gap by introducing \textbf{\model}, a large multi-modal model that can flexibly comprehend visual prompt inputs (\ie, points, boxes, and masks) and generate mask-grounded responses. Our model significantly differentiates itself from existing ones by unifying the internal representations of referred and segmented objects via a novel \textbf{object memory bank}, which is a hashmap storing the spatial-temporal information of object-of-interests. During inference, \model{} initializes the object memory bank and updates it on demand by adding object-centric information according to the context. The model responses are then generated conditioning on the fine-grained object memory. Benefits from such unification, \model{} is able to perform not only basic referring/segmentation tasks, but also flexible \textbf{pixel-level reasoning} tasks that require simultaneous visual prompt comprehension and mask prediction. As illustrated in Fig.~\ref{fig:task} (the last row), given a video\footnote{\textit{Images are treated as single-frame videos, thus we do not explicitly differentiate them in this work.}}, a question, and optionally a visual prompt (\eg, a point specified by a click on an object in any frame), \model{} can (1) infer the mask for the referred object in the corresponding frame, (2) propagate it to all video frames containing the same instance, (3) extract the mask-grounded object features, and finally (4) answer the question conditioning on both the video-level and object-centric information. All these operations are seamlessly conducted \textit{within a single model}, eliminating the need for external frame samplers \cite{visa}, mask generators \cite{osprey,videorefer}, or object trackers \cite{videolisa}.

\begin{figure}
\centering
\includegraphics[width=\linewidth]{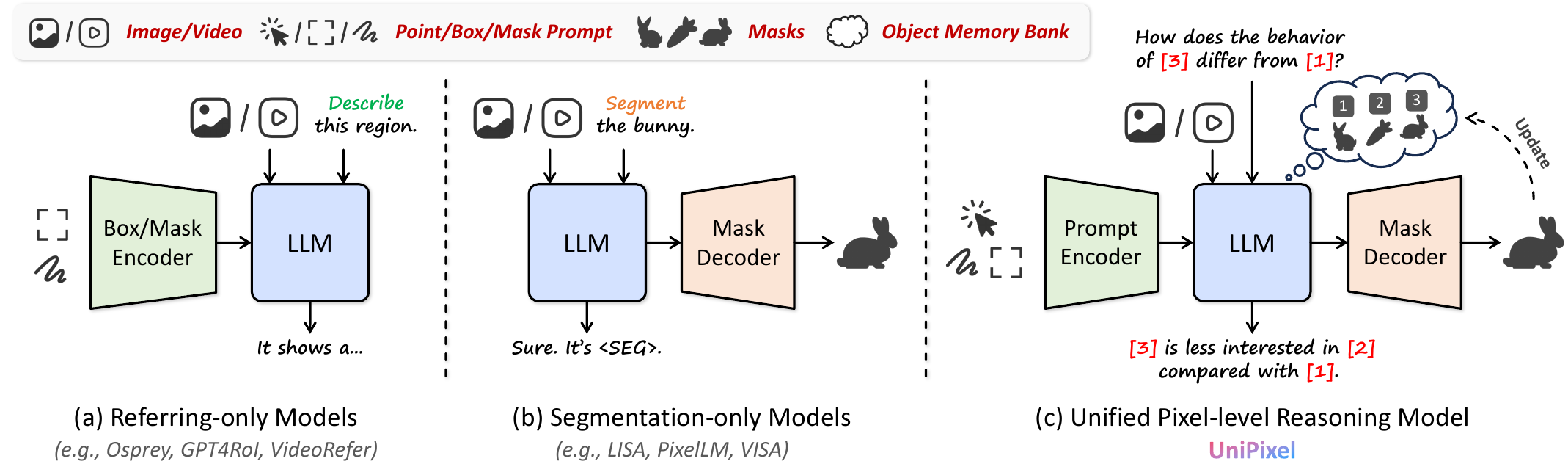}
\caption{\textbf{Schematic comparison between \model{} and its counterparts.} To the best of our knowledge, \model{} is the first unified method supporting simultaneous object referring and segmentation.}
\label{fig:compare}
\vspace{-5mm}
\end{figure}

We evaluate the effectiveness of \model{} from two aspects, \ie, basic referring/segmentation capabilities and flexible pixel-level reasoning capabilities. For the first aspect, we conduct extensive experiments on 10 public benchmarks across 8 image/video referring/segmentation tasks. Our method achieves state-of-the-art performance in diverse scenarios. Notably, on the challenging video reasoning segmentation and referred video QA tasks, our 3B model obtains \textbf{62.1} $\mathcal{J\&F}$ on ReVOS \cite{visa} and \textbf{72.8\%} Acc on VideoRefer-Bench$^{\rm Q}$ \cite{videorefer}, surpassing strong counterparts with 7B\,$\sim$\,13B parameters. Further ablation studies also demonstrate the mutual reinforcement effect of referring and segmentation. For the second aspect, we introduce a novel \textbf{PixelQA} task that jointly requires object-centric referring, segmentation, and QA in videos, which cannot be handled by existing methods. \model{} establishes a strong baseline for this novel setting. Our contributions are summarized below:

\begin{enumerate}[left=1.5em,topsep=0pt,itemsep=2pt,parsep=2pt]
\item We propose \textbf{\model}, a unified large multi-modal model that supports flexible object referring and segmentation in images and videos, via a novel \textbf{object memory bank} design.
\item Our model achieves state-of-the-art performance on 10 public benchmarks across 9 referring/segmentation tasks, verifying the \textbf{mutual reinforcement effect} of such unification.
\item We also introduce a novel \textbf{PixelQA} task that jointly requires object-centric referring, segmentation, and QA in videos, where \model{} establishes a strong baseline for this setting.
\end{enumerate}

\section{Related Work}

\textbf{Large Multi-modal Models}\quad The remarkable success of large multi-modal models (LMMs) has shifted the paradigm of visual-language understanding from close-ended experts to open-ended task solvers. Early attempts \cite{llava,llava1.5,instructblip,minigpt4} involve an MLP projector or Q-Former \cite{blip2} to align visual encoders to LLMs, enabling open-ended tasks such as visual question answering. With advanced designs such as dynamic resolution and data augmentation, open-source models, \eg, Qwen-VL \cite{qwenvl,qwen2vl,qwen2.5vl} and InternVL \cite{internvl,internvl2,internvl2.5} series, have narrowed the gap with advanced proprietary models like the GPT \cite{gpt4v,gpt4o} and Gemini families \cite{gemini1.5,gemini2.0}. Recent studies \cite{openaio1,deepseekr1,visionmath,starr1,videomind} also explore test-time scaling on visual-language understanding. However, these methods are spatially coarse-grained. \model{} can also be regarded as an object-centric test-time scaling approach, where key objects are first segmented then encoded to facilitate the subsequent reasoning process.

\textbf{Visual Referring and Segmentation}\quad To meet the growing demand for fine-grained visual understanding \cite{consnet,catnet,umt,r2tuning,vtgsurvey}, recent efforts have focused on enhancing LMMs with object referring and segmentation capabilities, as compared in Fig.~\ref{fig:compare}. LISA \cite{lisa} is a representative model that enables LMM-based segmentation by integrating SAM \cite{sam} as its decoder. They also introduced a novel reasoning segmentation task, requiring models to perform segmentation based on implicit queries. Other works in this direction \cite{nextchat,pixellm,groundhog,glamm,mmr} have explored advanced mask decoders, more flexible tasks, and larger-scale datasets. Recent studies have also extended these capabilities to videos \cite{videolisa,visa,sa2va}. Additionally, some research has examined regional understanding through boxes \cite{gpt4roi} and masks \cite{osprey,videorefer}. While recent approaches attempt to unify these two capabilities, they either support only images \cite{glamm} or rely on sub-optimal, tool-based pipelines \cite{vitron}. To the best of knowledge, \model{} is the first end-to-end method unifying object referring and mask prediction.

\begin{figure}
\centering
\includegraphics[width=0.95\linewidth]{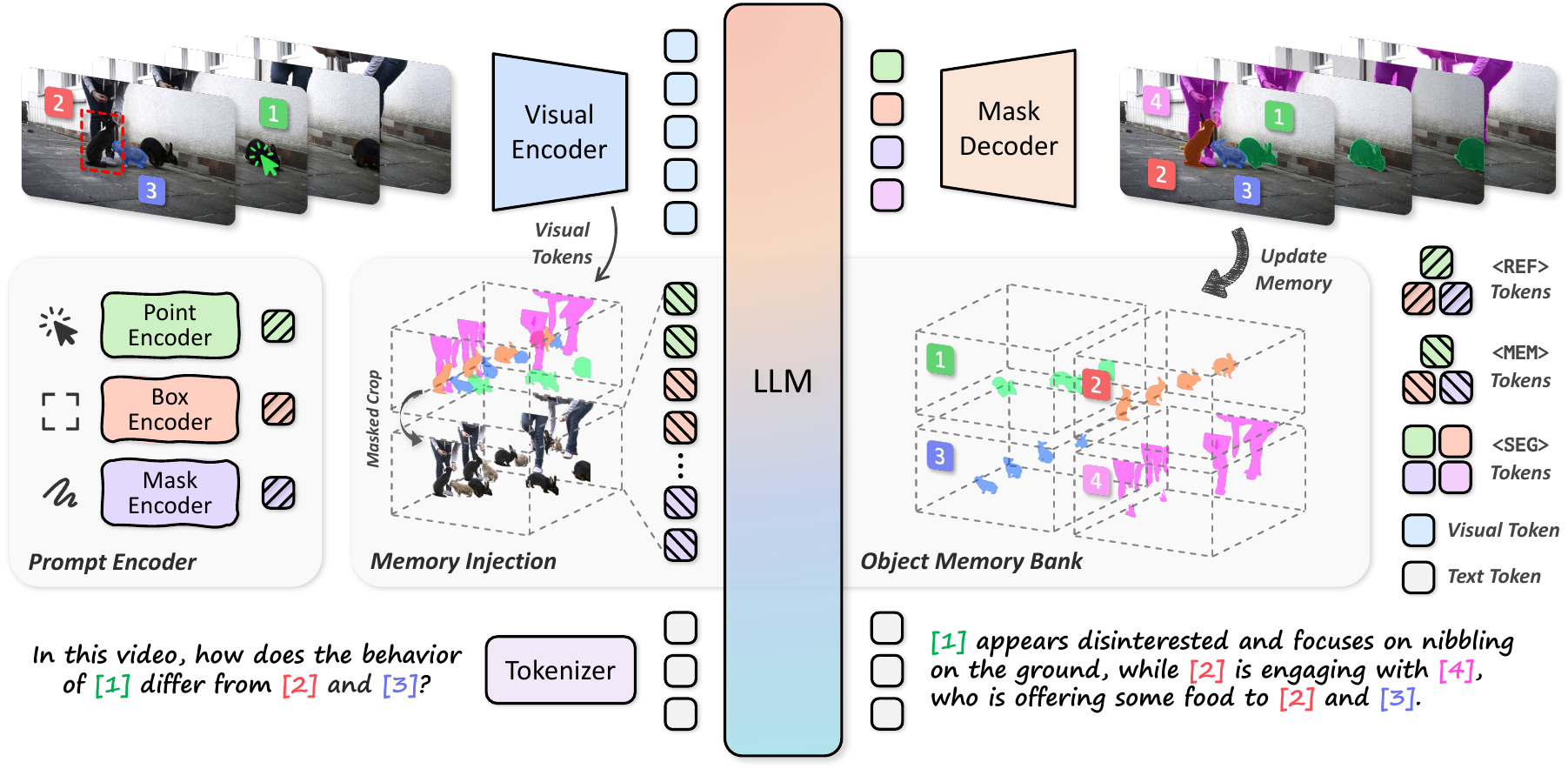}
\caption{\textbf{The architecture of \model{}.} Given a video, a question, and visual prompts, the model encodes them into tokens via the visual encoder, prompt encoder, and tokenizer, respectively, then predicts a spatial-temporal mask for each visual prompt via the mask decoder. The masks are updated into the object memory bank, and subsequently injected into the prompt for pixel-level reasoning.}
\label{fig:model}
\vspace{-3mm}
\end{figure}

\section{Method}

\textbf{Problem Formulation}\quad We provide a unified definition for pixel-level reasoning tasks. Formally, the inputs are an image or a video $\mathcal{X}$, a text prompt $\mathcal{T}$, and $N$ optional visual prompts $\{\mathcal{P}_i\}_{i=1}^{N}$ where each $\mathcal{P}_i$ could be a point, box, or mask on a specific frame. The outputs are textual responses to the prompt with $K$ grounded spatial-temporal masks $\{\mathcal{M}_i\}_{i=1}^{K}$. Here, both $N$ and $K$ could be zero (degenerating to a normal visual understanding task) and $K$ is not necessarily equal to $N$, as the model may segment extra objects/regions that are not specified by the visual prompts.

\textbf{Overview}\quad Fig.~\ref{fig:model} presents an overview of \model{}. It is built upon the Qwen2.5-VL \cite{qwen2.5vl} framework, consisting of an LLM backbone and a ViT-based visual encoder that supports dynamic resolution inputs. Given a video and a text prompt, the model first tokenizes them via the visual encoder and text tokenizer, then sends them into the LLM for response generation. To boost this framework from holistic-level to pixel-level, we introduce (1) a \textit{prompt encoder} (Sec.\ref{sec:prompt}) supporting three types of visual prompts, (2) an \textit{object memory bank} (Sec.\ref{sec:object}) for storing object information and injecting it into the response generation process, and (3) a \textit{mask decoder} (Sec.\ref{sec:mask}) for generating spatial-temporal masks. We also extend the LLM's vocabulary by adding \texttt{<REF>}, \texttt{<MEM>}, and \texttt{<SEG>} tokens. The former two serve as placeholders in the input prompt that would be replaced by visual prompt and memory tokens, respectively, while the \texttt{<SEG>} token is utilized to trigger and guide the mask decoding process. Detailed designs and interactions among these components are illustrated as follows.

\vspace{-1mm}
\subsection{Prompt Encoder}\label{sec:prompt}
\vspace{-1mm}

\begin{wrapfigure}{r}{0.53\textwidth}
\vspace{-4mm}
\includegraphics[width=\linewidth]{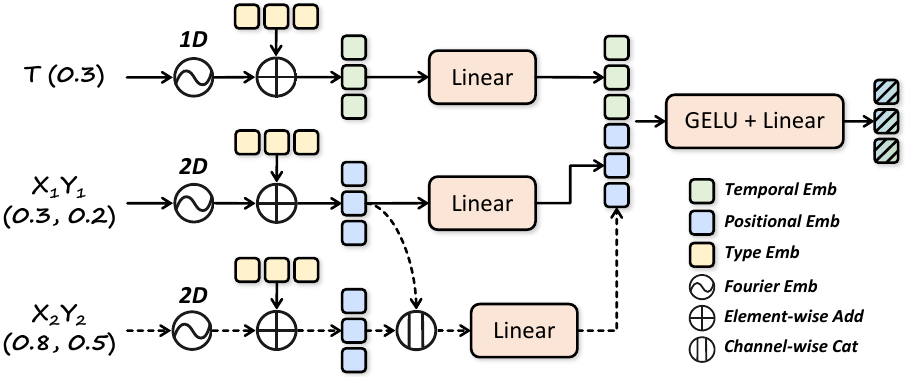}
\vspace{-4mm}
\caption{\textbf{Joint positional \& temporal encoding} for point ($X_1 Y_1 T$) and box ($X_1 Y_1 X_2 Y_2 T$) prompts.}
\label{fig:prompt}
\vspace{-4mm}
\end{wrapfigure}

This module aims to effectively encode each visual prompt into a single token that can be processed by the LLM. We denote a point prompt as a tuple $(x, y, t)$ containing its spatial coordinates $(x, y)$ and the corresponding frame index $t$. For box prompts, it is extended to $(x_1, y_1, x_2, y_2, t)$ containing the positions of top-left and bottom-right corners. A mask prompt is densely represented by a 2D binary mask $\mathbf{m}_{ij} \in \{0, 1\}$ with the same shape as the encoded target frame.

For sparse prompts (points and boxes), as shown in Fig.~\ref{fig:prompt}, we encode each position $(x_i, y_i)$ as the sum of a 2D Fourier embedding \cite{fourierembedding} and a learnable type embedding (indicating whether it is a single point, top-left corner, or bottom-right corner). For box prompts, we merge the two positional embeddings by concatenating them along the channel dimension and linearly projecting them back to the original size. Frame indices are also encoded similarly with 1D Fourier embeddings. The resulting positional and temporal embeddings are concatenated again, and then projected to the LLM's embedding space via a \texttt{GELU}\,$\to$\,\texttt{Linear} block, such that the sparse coordinates in a point/box are encoded into a compact high-dimensional token. This design is inspired by \cite{sam,sam2} with two key differences: (1) the spatial-only embeddings are extended to include temporal information, and (2) the negative points are discarded. For dense prompts (masks), we directly resize the binary masks and apply masked pooling on the outputs of the visual encoder. An M$\to$L projector (\texttt{Linear}\,$\to$\,\texttt{GELU}\,$\to$\,\texttt{Linear}) is leveraged to project the pooled visual features to the LLM's embedding space.

\vspace{-1mm}
\subsection{Object Memory Bank}\label{sec:object}
\vspace{-1mm}

Although sparse prompts contain rich positional and temporal information indicating the objects that users are referring to, it is still hard for the model to focus on these important regions. Previous studies \cite{gpt4roi,osprey,videorefer} also confirm that direct region cropping can generally provide better object awareness compared to positional pointers. To seamlessly integrate such a mechanism while preserving the flexibility of visual prompts (\eg, allow pointing on a single frame instead of drawing complete masks on all frames), we propose an object memory bank to bridge sparse visual prompts and dense object masks. This is a hashmap where the keys are object IDs and the values are the corresponding spatial-temporal masks. It is initialized as an empty storage for every chat session, and is dynamically updated on demand. We define two operations for the object memory bank, namely \textit{memory pre-filling} and \textit{memory injection}. Below is an example of memory-enhanced multi-round conversation.

\begin{tcolorbox}[boxrule=1pt, rounded corners, left=2mm, right=2mm, top=2mm, bottom=2mm]
\fontsize{7.5pt}{8pt}\selectfont
\textbf{Prompt 1:} How does the behavior of [1] \texttt{<REF>} differ from [2] \texttt{<REF>} and [3] \texttt{<REF>}?
\begin{tcolorbox}[title={\texttt{<REF>} detected, enhancing the prompt with object memory.}, boxrule=1pt, rounded corners, left=2mm, right=2mm, top=2mm, bottom=2mm]
\textbf{Memory Pre-filling Response:} \\
The relevant regions for this question are [1] \texttt{<SEG>} [2] \texttt{<SEG>} [3] \texttt{<SEG>} [4] \texttt{<SEG>}. {\color{blue}\;\;$\leftarrow$ \textit{4 objects saved into the memory}} \\\\
\textbf{Memory Injected Prompt:} \\
Here is a video with 4 frames denoted as <1> to <4>. The highlighted regions are as follows: \\
\mbox{[1]}: <1> \texttt{<MEM>} <2> \texttt{<MEM>} <3> \texttt{<MEM>} {\color{blue}\;\;$\leftarrow$ \textit{This object cannot be seen in the last frame}} \\
\mbox{[2]}: <1> \texttt{<MEM>} <2> \texttt{<MEM>} <3> \texttt{<MEM>} <4> \texttt{<MEM>} \\
\mbox{[3]}: <1> \texttt{<MEM>} <2> \texttt{<MEM>} <3> \texttt{<MEM>} <4> \texttt{<MEM>} \\
\mbox{[4]}: <1> \texttt{<MEM>} <2> \texttt{<MEM>} <3> \texttt{<MEM>} <4> \texttt{<MEM>} \\
How does the behavior of [1] differ from [2] and [3]?
\end{tcolorbox}
\textbf{Response 1:} [1] appears disinterested and focuses on nibbling on the ground, while [2] is engaging with [4], who is offering some food to [2] and [3]. \\
\textbf{Prompt 2:} What food is [4] offering? {\color{blue}\;\;$\leftarrow$ \textit{Users can directly refer to objects in the memory}} \\
\textbf{Response 2:} [4] is offering carrots.
\end{tcolorbox}

\textbf{Memory Pre-filling}\quad This operation is triggered upon the detection of \texttt{<REF>} tokens in the input prompt, aiming to thoroughly analyze the referred objects and predict their corresponding masks. In this stage, the model responds with object IDs and \texttt{<SEG>} tokens for the relevant objects according to the context, and predicts their spatial-temporal masks accordingly. These object-mask pairs are then saved into the object memory bank.

\textbf{Memory Injection}\quad We inject the features of the saved objects into the prompt to enhance object-awareness. Similar to the mask prompt encoder described in Sec.~\ref{sec:prompt}, each frame-level object mask is downsampled to match the resolution of visual tokens. We then apply masked pooling to aggregate object-centric features. Each frame-level mask is condensed into a single feature token, projected through the mask projector, and subsequently utilized to replace the corresponding \texttt{<MEM>} token in the memory-injected prompt. Through this \textit{pre-filling and injection} mechanism, object-centric information is effectively integrated into the model inference process.

\textbf{Why using object memory bank?}\quad An alternative is directly appending a \texttt{<SEG>} token to each \texttt{<REF>} token, followed by masked pooled features obtained during inference. However, we do not adopt this approach for two reasons: (1) During mask prediction, the \texttt{<SEG>} tokens, due to the unidirectional nature of causal self-attention, are unable to aggregate the full context of the prompt, thereby compromising the quality of predicted masks. (2) By utilizing the object memory bank, we can effectively decouple regional understanding and mask prediction, allowing each to benefit from referring and segmentation data during training, thus enhancing both capabilities.

\begin{table}[t]
\scriptsize
\centering
\caption{Comparison with state-of-the-art methods on ReVOS \cite{visa} \texttt{val} split. The best and second-best results are marked \textbf{bold} and \underline{underlined}, respectively.}
\vspace{2mm}
\begin{tabularx}{0.885\linewidth}{l|c|ccc|ccc|ccc|c}
\toprule
\multirow{2.6}{*}{\textbf{Method}} & \multirow{2.6}{*}{\textbf{Size}} & \multicolumn{3}{c|}{\textbf{Referring}} & \multicolumn{3}{c|}{\textbf{Reasoning}} & \multicolumn{3}{c|}{\textbf{Overall}} & \multirow{2.6}{*}{$\mathcal{R}$} \\
\cmidrule{3-5} \cmidrule{6-8} \cmidrule{9-11}
&& $\mathcal{J}$ & $\mathcal{F}$ & $\mathcal{J\&F}$ & $\mathcal{J}$ & $\mathcal{F}$ & $\mathcal{J\&F}$ & $\mathcal{J}$ & $\mathcal{F}$ & $\mathcal{J\&F}$ \\
\midrule
\rowcolor{gray!10}\multicolumn{12}{l}{\textit{\textcolor{gray}{Non-LLM-based Specialists}}\vspace{0.5mm}} \\
MTTR \cite{mttr} & -- & 29.8 & 30.2 & 30.0 & 20.4 & 21.5 & 21.0 & 25.1 & 25.9 & 25.5 & 5.6 \\
LMPM \cite{mevis} & -- & 29.0 & 39.1 & 34.1 & 13.3 & 24.3 & 18.8 & 21.2 & 31.7 & 26.4 & 3.2 \\
ReferFormer \cite{referformer} & -- & 31.2 & 34.3 & 32.7 & 21.3 & 25.6 & 23.4 & 26.2 & 29.9 & 28.1 & 8.8 \\
\midrule
\rowcolor{gray!10}\multicolumn{12}{l}{\textit{\textcolor{gray}{LLM-based Generalists}}\vspace{0.5mm}} \\
LISA \cite{lisa} & 13B & 45.2 & 47.9 & 46.6 & 34.3 & 39.1 & 36.7 & 39.8 & 43.5 & 41.6 & 8.6 \\
TrackGPT \cite{trackgpt} & 13B & 48.3 & 50.6 & 49.5 & 38.1 & 42.9 & 40.5 & 43.2 & 46.8 & 45.0 & 12.8 \\
VISA \cite{visa} & 13B & 55.6 & 59.1 & 57.4 & 42.0 & 46.7 & 44.3 & 48.8 & 52.9 & 50.9 & 14.5 \\
HyperSeg \cite{hyperseg} & 3B & 56.0 & 60.9 & 58.5 & 50.2 & 55.8 & 53.0 & 53.1 & 58.4 & 55.7 & -- \\
InstructSeg \cite{instructseg} & 3B & 54.8 & 59.2 & 57.0 & 49.2 & 54.7 & 51.9 & 52.0 & 56.9 & 54.5 & -- \\
GLUS \cite{glus} & 7B & 56.0 & 60.7 & 58.3 & 48.8 & 53.9 & 51.4 & 52.4 & 57.3 & 54.9 & 17.9 \\
ViLLa \cite{villa} & 6B & -- & -- & -- & -- & -- & -- & 54.9 & 59.1 & 57.0 & -- \\
Sa2VA \cite{sa2va} & 4B & -- & -- & -- & -- & -- & -- & -- & -- & 53.2 & -- \\
\midrule
\rowcolor{blue!7.5} \textbf{\model} (Ours) & 3B & \underline{62.3} & \underline{66.7} & \underline{64.5} & \underline{57.1} & \underline{62.1} & \underline{59.6} & \underline{59.7} & \underline{64.4} & \underline{62.1} & \underline{19.0} \\
\rowcolor{blue!7.5} \textbf{\model} (Ours) & 7B & \textbf{63.9} & \textbf{67.8} & \textbf{65.8} & \textbf{59.4} & \textbf{63.7} & \textbf{61.5} & \textbf{61.7} & \textbf{65.7} & \textbf{63.7} & \textbf{19.4} \\
\bottomrule
\end{tabularx}
\label{tab:revos}
\vspace{-2mm}
\end{table}

\vspace{-1mm}
\subsection{Mask Decoder}\label{sec:mask}
\vspace{-1mm}

We adopt SAM 2.1 \cite{sam2} as the mask decoder to disentangle the discrete language modeling and continuous mask prediction capabilities. For each \texttt{<SEG>} token, we extract its last-layer hidden states, downsample them via an L$\to$M projector (architecturally identical to the M$\to$L projector), and reshape them into two tokens. Using two tokens ensures better preservation of object information when downsampling from high- to low-dimensional channel space. These tokens prompt the mask decoder to predict the mask on the first frame, which is then propagated to the other frames.

\vspace{-1mm}
\subsection{Model Training}
\vspace{-1mm}

The training loss for \model{} is a linear combination of language modeling loss and mask decoding losses \cite{sam2}, including a focal loss and dice loss for mask prediction, a mean-absolute-error (MAE) loss for IoU prediction, and a cross-entropy loss for objectness prediction. The loss weights are set to 1, 100, 5, 5, and 5, respectively. We train the model through a three-stage progressive alignment recipe. The datasets are listed in Tab.~\ref{tab:training}. In the first stage, we pre-train the sparse prompt encoder using 851K regional captioning data. Then, we align the LLM and mask decoder by training the L$\to$M projector on 87K referring segmentation data. In the last stage, we further unfreeze the M$\to$L projector and mask decoder, and apply LoRA \cite{lora} on the visual encoder and LLM. The model is jointly trained on a large-scale corpus with around 2M samples for diverse tasks.

\section{Experiments}

We evaluate the effectiveness of \model{} by conducting extensive experiments across a diverse set of benchmarks. Specifically, we study the following research questions.
\begin{enumerate}[label=\textbf{Q\arabic*.},topsep=0pt,itemsep=0pt,parsep=0pt]
\item Whether \model{} is flexible and effective on basic image/video referring and segmentation tasks compared to the corresponding representative methods?
\item How does it perform on the more challenging PixelQA task, which requires joint referring, segmentation, and question answering in videos?
\item What effects does each architectural design contribute? More importantly, does the unified modeling of referring and segmentation lead to a mutual reinforcement effect?
\end{enumerate}

Detailed information about the benchmarks, evaluation metrics, implementation details, and more experimental results can be found in the appendix.

\begin{table}[t]
\scriptsize
\setlength{\tabcolsep}{5.4pt}
\centering
\caption{Comparison with state-of-the-art methods on referring video object segmentation (RVOS) and motion-grounded video reasoning datasets, including MeViS \cite{mevis} (\texttt{val}), Ref-YouTube-VOS \cite{urvos} (\texttt{val}), Ref-DAVIS17 \cite{davis2017} (\texttt{val}), and GroundMoRe \cite{groundmore} (\texttt{test}). The best and second-best results are marked \textbf{bold} and \underline{underlined}, respectively.}
\vspace{2mm}
\begin{tabularx}{\linewidth}{l|c|ccc|ccc|ccc|ccc}
\toprule
\multirow{2.6}{*}{\textbf{Method}} & \multirow{2.6}{*}{\textbf{Size}} & \multicolumn{3}{c|}{\textbf{MeViS}} & \multicolumn{3}{c|}{\textbf{Ref-YouTube-VOS}} & \multicolumn{3}{c|}{\textbf{Ref-DAVIS17}} & \multicolumn{3}{c}{\textbf{GroundMoRe}} \\
\cmidrule{3-5} \cmidrule{6-8} \cmidrule{9-11} \cmidrule{12-14}
&& $\mathcal{J}$ & $\mathcal{F}$ & $\mathcal{J\&F}$ & $\mathcal{J}$ & $\mathcal{F}$ & $\mathcal{J\&F}$ & $\mathcal{J}$ & $\mathcal{F}$ & $\mathcal{J\&F}$ & $\mathcal{J}$ & $\mathcal{F}$ & $\mathcal{J\&F}$ \\
\midrule
\rowcolor{gray!10}\multicolumn{14}{l}{\textit{\textcolor{gray}{Non-LLM-based Specialists}}\vspace{0.5mm}} \\
ReferFormer \cite{referformer} & -- & 29.8 & 32.2 & 31.0 & 61.3 & 64.6 & 62.9 & 58.1 & 64.1 & 61.1 & 11.2 & 14.3 & 12.7 \\
LMPM \cite{mevis} & -- & 34.2 & 40.2 & 37.2 & -- & -- & -- & -- & -- & -- & 12.7 & 14.0 & 13.3 \\
OnlineRefer \cite{onlinerefer} & -- & -- & -- & -- & 61.6 & 65.5 & 63.5 & 61.6 & 67.7 & 64.8 & -- & -- & -- \\
\midrule
\rowcolor{gray!10}\multicolumn{14}{l}{\textit{\textcolor{gray}{LLM-based Generalists}}\vspace{0.5mm}} \\
PixelLM \cite{pixellm} & 7B & 36.3 & 41.1 & 38.7 & 54.3 & 55.7 & 55.0 & 63.4 & 70.0 & 66.7 & 9.9 & 10.0 & 10.0 \\
LISA \cite{lisa} & 13B & 35.8 & 40.0 & 37.9 & 54.0 & 54.8 & 54.4 & 63.2 & 68.8 & 66.0 & 6.3 & 6.7 & 6.5 \\
VISA \cite{visa} & 13B & 41.8 & 47.1 & 44.5 & 61.4 & 64.7 & 63.0 & 67.0 & 73.8 & 70.4 & 5.3 & 4.7 & 5.9 \\
VideoLISA \cite{videolisa} & 3.8B & 41.3 & 47.6 & 44.4 & 61.7 & 65.7 & 63.7 & 64.9 & 72.7 & 68.8 & -- & -- & -- \\
VideoGLaMM \cite{videoglamm} & 3.8B & 42.1 & 48.2 & 45.2 & 65.4 & 68.2 & 66.8 & 73.3 & 65.6 & 69.5 & -- & -- & -- \\
ViLLa \cite{villa} & 6B & 46.5 & 52.3 & 49.4 & 64.6 & 70.4 & 67.5 & 70.6 & 78.0 & 74.3 & -- & -- & -- \\
GLUS \cite{glus} & 7B & 48.5 & 54.2 & 51.3 & 65.5 & 69.0 & 67.3 & -- & -- & -- & -- & -- & -- \\
Sa2VA \cite{sa2va} & 4B & -- & -- & 46.2 & -- & -- & 70.0 & -- & -- & 73.8 & -- & -- & -- \\
MoRA \cite{groundmore} & 7B & -- & -- & -- & -- & -- & -- & -- & -- & -- & 27.4 & 26.9 & 27.2 \\
\midrule
\rowcolor{blue!7.5} \textbf{\model} (Ours) & 3B & \underline{50.4} & \underline{55.7} & \underline{53.1} & \underline{68.6} & \underline{72.3} & \underline{70.5} & \underline{70.7} & \underline{77.8} & \underline{74.2} & \underline{36.0} & \underline{38.7} & \underline{37.4} \\
\rowcolor{blue!7.5} \textbf{\model} (Ours) & 7B & \textbf{53.2} & \textbf{58.3} & \textbf{55.8} & \textbf{69.5} & \textbf{72.4} & \textbf{71.0} & \textbf{72.7} & \textbf{80.1} & \textbf{76.4} & \textbf{36.5} & \textbf{39.1} & \textbf{37.8} \\
\bottomrule
\end{tabularx}
\label{tab:rvos}
\vspace{-6mm}
\end{table}

\begin{table}[t]
\scriptsize
\centering
\caption{Comparison with state-of-the-art methods on image referring expression segmentation (RES) and reasoning segmentation datasets, including RefCOCO/+/g \cite{refcoco,refcocog} and ReasonSeg \cite{lisa} (\texttt{val}). The best and second-best results are marked \textbf{bold} and \underline{underlined}, respectively.}
\vspace{2mm}
\begin{tabularx}{0.899\linewidth}{l|c|ccc|ccc|cc|cc}
\toprule
\multirow{2.6}{*}{\textbf{Method}} & \multirow{2.6}{*}{\textbf{Size}} & \multicolumn{3}{c|}{\textbf{RefCOCO}} & \multicolumn{3}{c|}{\textbf{RefCOCO+}} & \multicolumn{2}{c|}{\textbf{RefCOCOg}} & \multicolumn{2}{c}{\textbf{ReasonSeg}} \\
\cmidrule{3-5} \cmidrule{6-8} \cmidrule{9-10} \cmidrule{11-12}
&& val & testA & testB & val & testA & testB & val(U) & test(U) & gIoU & cIoU \\
\midrule
\rowcolor{gray!10}\multicolumn{12}{l}{\textit{\textcolor{gray}{Non-LLM-based Specialists}}\vspace{0.5mm}} \\
ReLA \cite{gres} & -- & 73.8 & 76.5 & 70.2 & 66.0 & 71.0 & 57.7 & 65.0 & 66.0 & -- & -- \\
X-Decoder \cite{xdecoder} & -- & -- & -- & -- & -- & -- & -- & 64.6 & -- & 22.6 & 17.9 \\
SEEM \cite{seem} & -- & -- & -- & -- & -- & -- & -- & 65.7 & -- & 25.5 & 21.2 \\
\midrule
\rowcolor{gray!10}\multicolumn{12}{l}{\textit{\textcolor{gray}{LLM-based Image Generalists}}\vspace{0.5mm}} \\
NExT-Chat \cite{nextchat} & 7B & 74.7 & 78.9 & 69.5 & 65.1 & 71.9 & 56.7 & 67.0 & 67.0 & -- & -- \\
PixelLM \cite{pixellm} & 7B & 73.0 & 76.5 & 68.2 & 66.3 & 71.7 & 58.3 & 69.3 & 70.5 & -- & -- \\
LISA \cite{lisa} & 7B & 74.9 & 79.1 & 72.3 & 65.1 & 70.8 & 58.1 & 67.9 & 70.6 & 61.3 & \underline{62.9} \\
Groundhog \cite{groundhog} & 7B & 78.5 & 79.9 & 75.7 & 70.5 & 75.0 & 64.9 & 74.1 & 74.6 & 56.2 & -- \\
LaSagnA \cite{lasagna} & 7B & 76.8 & 78.7 & 73.8 & 66.4 & 70.6 & 60.1 & 70.6 & 71.9 & 48.8 & 47.2 \\
M$^{\rm 2}$SA \cite{mmr} & 13B & 74.6 & 77.6 & 71.0 & 64.0 & 68.1 & 57.6 & 69.0 & 69.3 & -- & -- \\
\midrule
\rowcolor{gray!10}\multicolumn{12}{l}{\textit{\textcolor{gray}{LLM-based Video Generalists}}\vspace{0.5mm}} \\
VideoLISA \cite{videolisa} & 3.8B & 73.8 & 76.6 & 68.8 & 63.4 & 68.8 & 56.2 & 68.3 & 68.8 & \underline{61.4} & \textbf{67.1} \\
VISA \cite{visa} & 7B & 72.4 & 75.5 & 68.1 & 59.8 & 64.8 & 53.1 & 65.5 & 66.4 & 52.7 & 57.8 \\
Vitron \cite{vitron} & 7B & 75.5 & 79.5 & 72.2 & 66.7 & 72.5 & 58.0 & 67.9 & 68.9 & -- & -- \\
Sa2VA \cite{sa2va} & 4B & 78.9 & -- & -- & 71.7 & -- & -- & 74.1 & -- & -- & -- \\
\midrule
\rowcolor{blue!7.5} \textbf{\model} (Ours) & 3B & \underline{80.5} & \underline{82.6} & \underline{76.9} & \underline{74.3} & \underline{78.9} & \underline{68.4} & \underline{76.3} & \underline{77.0} & \textbf{64.0} & 56.2 \\
\rowcolor{blue!7.5} \textbf{\model} (Ours) & 7B & \textbf{80.8} & \textbf{83.0} & \textbf{77.4} & \textbf{75.3} & \textbf{80.1} & \textbf{70.0} & \textbf{76.4} & \textbf{77.1} & 60.5 & 58.7 \\
\bottomrule
\end{tabularx}
\label{tab:res}
\vspace{-6mm}
\end{table}

\subsection{Q1: Comparison with State-of-the-Arts on Referring and Segmentation Tasks}\label{sec:sota}

\textbf{Reasoning Video Object Segmentation}\quad We begin with the most challenging ReVOS \cite{visa} dataset, which requires models to predict masks based on implicit text queries demanding complex reasoning abilities based on world knowledge. The results are shown in Tab.~\ref{tab:revos}. Our 3B variant outperforms all existing methods with larger LLMs (including Sa2VA-4B \cite{sa2va} also with SAM 2 decoder), achieving 62.1 overall $\mathcal{J\&F}$. The 7B model further boosts the performance to 64.0 $\mathcal{J\&F}$ -- an improvement of 12\% over the previous state-of-the-art -- demonstrating that \model{} can effectively understand implicit queries based on its world knowledge, and accurately generate masks as responses.

\begin{table}
\scriptsize
\begin{minipage}{0.5\textwidth}
\caption{Experimental results on MeViS \cite{mevis} \texttt{val$^\texttt{u}$} set. \underline{Post} means applying post optimization.}
\vspace{2mm}
\setlength{\tabcolsep}{5.4pt}
\begin{tabularx}{\linewidth}{p{2.4cm}<{\raggedright}|cc|ccc}
\toprule
\textbf{Method} & \textbf{Size} & \textbf{FT} & $\mathcal{J}$ & $\mathcal{F}$ & $\mathcal{J\&F}$ \\
\midrule
LMPM \cite{mevis} & -- & \xmark & 36.5 & 43.9 & 40.2 \\
\midrule
LISA \cite{lisa} & 7B & \xmark & 39.9 & 46.5 & 43.2 \\
LISA \cite{lisa} + XMem \cite{xmem} & 7B & \xmark & 41.9 & 49.3 & 45.6 \\
VideoLISA \cite{videolisa} & 7B & \xmark & 48.4 & 54.9 & 51.7 \\
VideoLISA \cite{videolisa} + Post & 7B & \xmark & 50.9 & 58.1 & 54.5 \\
Sa2VA \cite{sa2va} & 4B & \xmark & -- & -- & 52.1 \\
Sa2VA \cite{sa2va} & 8B & \xmark & -- & -- & 57.0 \\
\midrule
\rowcolor{blue!7.5} \textbf{\model} (Ours) & 3B & \xmark & \underline{56.1} & \underline{63.2} & \underline{59.7} \\
\rowcolor{blue!7.5} \textbf{\model} (Ours) & 7B & \xmark & \textbf{58.4} & \textbf{65.0} & \textbf{61.7} \\
\bottomrule
\end{tabularx}
\label{tab:mevis_valid_u}
\end{minipage}
\hfill
\begin{minipage}{0.48\textwidth}
\caption{Comparison on Ref-SAV \cite{sa2va} \texttt{val} set. \underline{FT} means fine-tuning after pre-/co-training.}
\vspace{2mm}
\begin{tabularx}{\linewidth}{p{1.85cm}<{\raggedright}|cc|ccc}
\toprule
\textbf{Method} & \textbf{Size} & \textbf{FT} & $\mathcal{J}$ & $\mathcal{F}$ & $\mathcal{J\&F}$ \\
\midrule
UniRef++ \cite{uniref} & -- & \xmark & 11.6 & 9.5 & 10.5 \\
UNINEXT \cite{uninext} & -- & \xmark & 8.8 & 6.4 & 7.6 \\
LMPM \cite{mevis} & -- & \xmark & 12.2 & 9.8 & 10.3 \\
VISA \cite{visa} & 7B & \xmark & 13.2 & 11.3 & 11.8 \\
Sa2VA \cite{sa2va} & 8B & \xmark & 39.6 & 43.0 & 41.3 \\
\midrule
\textcolor{gray}{UniRef++ \cite{uniref}} & \textcolor{gray}{--} & \textcolor{gray}{\cmark} & \textcolor{gray}{15.8} & \textcolor{gray}{13.4} & \textcolor{gray}{14.6} \\
\textcolor{gray}{Sa2VA \cite{sa2va}} & \textcolor{gray}{8B} & \textcolor{gray}{\cmark} & \textcolor{gray}{48.3} & \textcolor{gray}{51.7} & \textcolor{gray}{50.0} \\
\midrule
\rowcolor{blue!7.5} \textbf{\model} (Ours) & 3B & \xmark & \underline{66.9} & \underline{67.6} & \underline{67.2} \\
\rowcolor{blue!7.5} \textbf{\model} (Ours) & 7B & \xmark & \textbf{68.5} & \textbf{69.6} & \textbf{69.0} \\
\bottomrule
\end{tabularx}
\label{tab:ref_sav}
\end{minipage}
\vspace{-5mm}
\end{table}

\begin{table}
\scriptsize
\centering
\caption{Fine-tuned performance on referring expression segmentation (RES) datasets, including RefCOCO/+/g \cite{refcoco,refcocog}. The best and second-best results are marked \textbf{bold} and \underline{underlined}, respectively.}
\vspace{2mm}
\begin{tabularx}{0.7815\linewidth}{l|c|ccc|ccc|cc}
\toprule
\multirow{2.6}{*}{\textbf{Method}} & \multirow{2.6}{*}{\textbf{Size}} & \multicolumn{3}{c|}{\textbf{RefCOCO}} & \multicolumn{3}{c|}{\textbf{RefCOCO+}} & \multicolumn{2}{c}{\textbf{RefCOCOg}} \\
\cmidrule{3-5} \cmidrule{6-8} \cmidrule{9-10}
&& val & testA & testB & val & testA & testB & val(U) & test(U) \\
\midrule
LISA \cite{lisa} & 7B & 74.9 & 79.1 & 72.3 & 65.1 & 70.8 & 58.1 & 67.9 & 70.6 \\
GSVA \cite{gsva} & 7B & 77.2 & 78.9 & 73.5 & 65.9 & 69.6 & 59.8 & 72.7 & 73.3 \\
OMG-LLaVA \cite{omgllava} & 7B & 78.0 & 80.3 & 74.1 & 69.1 & 73.1 & 63.0 & 72.9 & 72.9 \\
GLaMM \cite{glamm} & 7B & 79.5 & 83.2 & 76.9 & 72.6 & 78.7 & 64.6 & 74.2 & 74.9 \\
Sa2VA \cite{sa2va} & 4B & 80.4 & -- & -- & 74.3 & -- & -- & 75.7 & -- \\
\midrule
\rowcolor{blue!7.5} \textbf{\model} (Ours) & 3B & \underline{81.9} & \underline{83.5} & \underline{78.6} & \underline{75.3} & \underline{80.3} & \underline{70.6} & \underline{77.2} & \underline{78.5} \\
\rowcolor{blue!7.5} \textbf{\model} (Ours) & 7B & \textbf{83.0} & \textbf{84.9} & \textbf{80.4} & \textbf{77.8} & \textbf{82.3} & \textbf{72.7} & \textbf{78.7} & \textbf{79.7} \\
\bottomrule
\end{tabularx}
\label{tab:res_ft}
\vspace{-5mm}
\end{table}

\begin{table}[h!]
\scriptsize
\centering
\caption{Experimental results on referring expression comprehension (REC) datasets, including RefCOCO/+/g \cite{refcoco,refcocog}. The best and second-best results are marked \textbf{bold} and \underline{underlined}, respectively.}
\vspace{2mm}
\begin{tabularx}{0.7535\linewidth}{l|c|ccc|ccc|cc}
\toprule
\multirow{2.6}{*}{\textbf{Method}} & \multirow{2.6}{*}{\textbf{Size}} & \multicolumn{3}{c|}{\textbf{RefCOCO}} & \multicolumn{3}{c|}{\textbf{RefCOCO+}} & \multicolumn{2}{c}{\textbf{RefCOCOg}} \\
\cmidrule{3-5} \cmidrule{6-8} \cmidrule{9-10}
&& val & testA & testB & val & testA & testB & val(U) & test(U) \\
\midrule
OFA \cite{ofa} & -- & 80.0 & 83.7 & 76.4 & 68.3 & 76.0 & 61.8 & 67.6 & 67.6 \\
Shikra \cite{shikra} & 7B & 87.0 & 90.6 & 80.2 & 81.6 & 87.4 & 72.1 & 82.3 & 82.2 \\
MiniGPT-v2 \cite{minigptv2} & 7B & 88.7 & 91.6 & 85.3 & 79.9 & 85.1 & 74.4 & 84.4 & 84.6 \\
Vitron \cite{vitron} & 7B & 90.9 & 93.2 & \textbf{89.3} & 83.7 & 89.1 & 76.9 & 86.4 & 87.0 \\
\midrule
\rowcolor{blue!7.5} \textbf{\model} (Ours) & 3B & \underline{91.8} & \underline{93.8} & 87.5 & \underline{86.3} & \underline{90.8} & \underline{80.3} & \underline{88.0} & \underline{88.2} \\
\rowcolor{blue!7.5} \textbf{\model} (Ours) & 7B & \textbf{92.0} & \textbf{94.4} & \underline{88.1} & \textbf{87.2} & \textbf{91.9} & \textbf{82.1} & \textbf{88.6} & \textbf{88.7} \\
\bottomrule
\end{tabularx}
\label{tab:rec}
\vspace{-5mm}
\end{table}

\textbf{Referring Video Object Segmentation}\quad The performance comparisons on MeViS \cite{mevis}, Ref-YouTube-VOS \cite{urvos}, and Ref-DAVIS17 \cite{davis2017} datasets are presented in Tab.~\ref{tab:rvos}. \model{} consistently achieves the best performance among its counterparts. Its advantage is particularly evident on the more challenging MeViS dataset, where our 3B model outperforms GLUS-7B \cite{glus} by around 3.5\%, as well as the similarly sized VideoGLaMM-3.8B \cite{videoglamm} by 17\%. More experimental results on MeViS \cite{mevis} \texttt{val$^\texttt{u}$} set and Ref-SAV \cite{sa2va} \texttt{val} set are provided in Tab.~\ref{tab:mevis_valid_u} and Tab.~\ref{tab:ref_sav}, respectively. Ref-SAV features long referring descriptions, large object motion, large camera motion, and heavy occlusion compared with existing datasets. Given these complex descriptions and video content, our method consistently performs better than counterparts, including those fine-tuned on the target dataset.

\textbf{Motion-Grounded Video Reasoning}\quad We also evaluate our method on GroundMoRe \cite{groundmore} dataset (results shown in Tab.~\ref{tab:rvos}), which highlights visual answer generation that requires joint spatial and temporal grounding. Note that we mainly compare the results with MoRA \cite{groundmore}, which is fine-tuned on GroundMoRe while other methods are evaluated under the zero-shot setting. Benefit from the strong pixel-level reasoning capability, \model{} significantly performs better than the baseline.

\textbf{Referring Expression Segmentation and Reasoning Segmentation}\quad Tab.~\ref{tab:res} compares the image segmentation capabilities using explicit and implicit queries. We evaluate our co-trained model on RefCOCO/+/g \cite{refcoco,refcocog} and ReasonSeg \cite{lisa}. While state-of-the-art performance has been achieved on RES datasets, we observe that the reasoning segmentation data (239 samples) can be easily overwhelmed by the other samples during training due to its limited size. Tab.~\ref{tab:res_ft} presents the RES performance after fine-tuning. We follow the common practice that jointly fine-tunes the model on RefCOCO/+/g datasets \cite{refcoco,refcocog}, and then evaluate on them separately. These results demonstrate the generalizability of \model{} when facing both explicit and implicit queries.

\textbf{Referring Expression Comprehension}\quad Our method also supports referring expression comprehension by inferring the bounding boxes from predicted masks. Its performance (accuracy with IoU $\geqslant$ 0.5) is compared with representative methods in Tab.~\ref{tab:rec}. Benefiting from the high-quality mask prediction, \model{} can also achieve very competitive performance on this simpler task.

\textbf{Referred Video Description and Question Answering}\quad We study \model{}'s regional understanding capabilities on VideoRefer-Bench \cite{videorefer}, which contains two subsets for description and question answering tasks. The comparisons are in Tab.~\ref{tab:videoreferd} and Tab.~\ref{tab:videoreferq}. BQ, SQ, RQ, CQ, and FP denote basic questions, sequential questions, relational questions, reasoning questions, and future predictions, respectively. Both tasks leverage mask prompts as inputs, where single-frame and multi-frame modes denote applying the masks only on a specific frame and on all frames, respectively. \model{} can effectively comprehend both types of prompts, and accurately respond with object-centric descriptions or answers, surpassing strong models including GPT-4o \cite{gpt4o} and VideoRefer \cite{videorefer}.

\begin{table}[t]
\scriptsize
\centering
\caption{Comparison with state-of-the-art methods on VideoRefer-Bench$^{\rm D}$ \cite{videorefer}. The best and second-best results are marked \textbf{bold} and \underline{underlined}, respectively.}
\vspace{2mm}
\begin{tabularx}{0.835\linewidth}{l|c|ccccc|ccccc}
\toprule
\multirow{2.6}{*}{\textbf{Method}} & \multirow{2.6}{*}{\textbf{Size}} & \multicolumn{5}{c|}{\textbf{Single-Frame}} & \multicolumn{5}{c}{\textbf{Multi-Frame}} \\
\cmidrule{3-7} \cmidrule{8-12}
&& \textbf{SC} & \textbf{AD} & \textbf{TD} & \textbf{HD} & \textbf{Avg.} & \textbf{SC} & \textbf{AD} & \textbf{TD} & \textbf{HD} & \textbf{Avg.} \\
\midrule
\rowcolor{gray!10}\multicolumn{12}{l}{\textit{\textcolor{gray}{General LMMs}}\vspace{0.5mm}} \\
LLaVA-OV \cite{llavaonevision} & 7B & 2.62 & 1.58 & 2.19 & 2.07 & 2.12 & 3.09 & 1.94 & 2.50 & 2.41 & 2.48 \\
Qwen2-VL \cite{qwen2vl} & 7B & 2.97 & 2.24 & 2.03 & 2.31 & 2.39 & 3.30 & 2.54 & 2.22 & 2.12 & 2.55 \\
InternVL2 \cite{internvl2} & 26B & 3.55 & 2.99 & 2.57 & 2.25 & 2.84 & \underline{4.08} & \textbf{3.35} & 3.08 & 2.28 & 3.20 \\
GPT-4o-mini \cite{gpt4o} & -- & 3.56 & 2.85 & 2.87 & 2.38 & 2.92 & 3.89 & 3.18 & 2.62 & 2.50 & 3.05 \\
GPT-4o \cite{gpt4o} & -- & 3.34 & 2.96 & 3.01 & 2.50 & 2.95 & 4.15 & \underline{3.31} & \underline{3.11} & 2.43 & 3.25 \\
\midrule
\rowcolor{gray!10}\multicolumn{12}{l}{\textit{\textcolor{gray}{Image Referring LMMs}}\vspace{0.5mm}} \\
Ferret \cite{ferret} & 7B & 3.08 & 2.01 & 1.54 & 2.14 & 2.19 & 3.20 & 2.38 & 1.97 & 1.38 & 2.23 \\
Osprey \cite{osprey} & 7B & 3.19 & 2.16 & 1.54 & 2.45 & 2.34 & 3.30 & 2.66 & 2.10 & 1.58 & 2.41 \\
\midrule
\rowcolor{gray!10}\multicolumn{12}{l}{\textit{\textcolor{gray}{Video Referring LMMs}}\vspace{0.5mm}} \\
Elysium \cite{elysium} & 7B & 2.35 & 0.30 & 0.02 & \underline{3.59} & 1.57 & -- & -- & -- & -- & -- \\
Artemis \cite{artemis} & 7B & -- & -- & -- & -- & -- & 3.42 & 1.34 & 1.39 & 2.90 & 2.26 \\
VideoRefer \cite{videorefer}	 & 7B & \textbf{4.41} & \textbf{3.27} & \underline{3.03} & 2.97 & \textbf{3.42} & \textbf{4.44} & 3.27 & 3.10 & 3.04 & \textbf{3.46} \\
\midrule
\rowcolor{blue!7.5} \textbf{\model} (Ours) & 3B & \underline{4.04} & \underline{3.15} & \textbf{3.10} & 3.37 & \textbf{3.42} & \underline{4.08} & 3.13 & \textbf{3.13} & \underline{3.42} & \underline{3.44} \\
\rowcolor{blue!7.5} \textbf{\model} (Ours) & 7B & 3.83 & 3.07 & 2.96 & \textbf{3.62} & \underline{3.37} & 3.82 & 3.05 & 3.01 & \textbf{3.57} & 3.36 \\
\bottomrule
\end{tabularx}
\label{tab:videoreferd}
\vspace{-3mm}
\end{table}

\begin{table}
\scriptsize
\setlength{\tabcolsep}{4pt}
\begin{minipage}{0.555\textwidth}
\caption{Comparison with state-of-the-art methods on VideoRefer-Bench$^{\rm Q}$ \cite{videorefer} (\textit{mask prompts}). \underline{MF} denotes multi-frame mode. Full question types are in Sec.~\ref{sec:sota}.}
\vspace{2mm}
\begin{tabularx}{\linewidth}{l|cc|cccccc}
\toprule
\textbf{Method} & \textbf{Size} & \textbf{MF} & \textbf{BQ} & \textbf{SQ} & \textbf{RQ} & \textbf{CQ} & \textbf{FP} & \textbf{Avg.} \\
\midrule
\rowcolor{gray!10}\multicolumn{9}{l}{\textit{\textcolor{gray}{General LMMs}}\vspace{0.5mm}} \\
LLaVA-OV \cite{llavaonevision} & 7B & \xmark{} & 58.7 & 62.9 & 64.7 & 87.4 & 76.3 & 67.4 \\
Qwen2-VL \cite{qwen2vl} & 7B & \xmark{} & 62.0 & 69.6 & 54.9 & 87.3 & 74.6 & 66.0 \\
InternVL2 \cite{internvl2} & 26B & \xmark{} & 58.5 & 63.5 & 53.4 & 88.0 & \underline{78.9} & 65.0 \\
GPT-4o-mini \cite{gpt4o} & -- & \xmark{} & 57.6 & 67.1 & 56.5 & 85.9 & 75.4 & 65.8 \\
GPT-4o \cite{gpt4o} & -- & \xmark{} & 62.3 & \textbf{74.5} & \textbf{66.0} & 88.0 & 73.7 & 71.3 \\
\midrule
\rowcolor{gray!10}\multicolumn{9}{l}{\textit{\textcolor{gray}{Image Referring LMMs}}\vspace{0.5mm}} \\
Ferret \cite{ferret} & 7B & \xmark{} & 35.2 & 44.7 & 41.9 & 70.4 & 74.6 & 48.8 \\
Osprey \cite{osprey} & 7B & \xmark{} & 45.9 & 47.1 & 30.0 & 48.6 & 23.7 & 39.9 \\
\midrule
\rowcolor{gray!10}\multicolumn{9}{l}{\textit{\textcolor{gray}{Video Referring LMMs}}\vspace{0.5mm}} \\
VideoRefer \cite{videorefer} & 7B & \xmark{} & \textbf{75.4} & 68.6 & 59.3 & \textbf{89.4} & 78.1 & 71.9 \\
\rowcolor{blue!7.5} \textbf{\model} (Ours) & 3B & \xmark{} & \underline{73.6} & 70.3 & 60.7 & \underline{88.8} & 78.0 & \underline{72.2} \\
\rowcolor{blue!7.5} \textbf{\model} (Ours) & 7B & \xmark{} & 68.9 & \underline{73.1} & \underline{64.7} & \underline{88.8} & \textbf{83.3} & \textbf{73.4} \\
\midrule
VideoRefer \cite{videorefer} & 7B & \cmark{} & -- & 70.6 & 60.5 & -- & -- & 72.1 \\
\rowcolor{blue!7.5} \textbf{\model} (Ours) & 3B & \cmark{} & \textbf{75.3} & \underline{70.7} & \underline{62.3} & \underline{87.4} & \underline{77.2} & \underline{72.8} \\
\rowcolor{blue!7.5} \textbf{\model} (Ours) & 7B & \cmark{} & \underline{70.6} & \textbf{74.6} & \textbf{64.7} & \textbf{88.8} & \textbf{82.5} & \textbf{74.1} \\
\bottomrule
\end{tabularx}
\label{tab:videoreferq}
\end{minipage}
\hfill
\begin{minipage}{0.41\textwidth}
\caption{Evaluation results on our newly introduced PixelQA task. All the visual prompts are applied in a single frame. See Sec.~\ref{sec:pixelqa} for detailed settings.}
\vspace{2mm}
\begin{tabularx}{\linewidth}{l|c|ccc|c}
\toprule
\textbf{Method} & \textbf{Size} & $\mathcal{J}$ & $\mathcal{F}$ & $\mathcal{J\&F}$ & \textbf{Acc} \\
\midrule
\rowcolor{gray!10}\multicolumn{6}{l}{\textit{\textcolor{gray}{Point Prompts}}\vspace{0.73mm}} \\
InternVL2 \cite{internvl2} & 26B & -- & -- & -- & 60.8 \\
Qwen2-VL \cite{qwen2vl} & 72B & -- & -- & -- & 69.3 \\
\rowcolor{blue!7.5} \textbf{\model} (Ours) & 3B & \textbf{57.3} & \textbf{64.4} & \textbf{60.9} & \underline{71.1} \\
\rowcolor{blue!7.5} \textbf{\model} (Ours) & 7B & \underline{42.1} & \underline{47.1} & \underline{44.6} & \textbf{71.4} \\
\midrule
\rowcolor{gray!10}\multicolumn{6}{l}{\textit{\textcolor{gray}{Box Prompts}}\vspace{0.73mm}} \\
InternVL2 \cite{internvl2} & 26B & -- & -- & -- & 61.3 \\
Qwen2-VL \cite{qwen2vl} & 72B & -- & -- & -- & 69.0 \\
\rowcolor{blue!7.5} \textbf{\model} (Ours) & 3B & \textbf{57.8} & \textbf{64.7} & \textbf{61.3} & \underline{70.3} \\
\rowcolor{blue!7.5} \textbf{\model} (Ours) & 7B & \underline{41.1} & \underline{46.4} & \underline{43.8} & \textbf{71.4} \\
\midrule
\rowcolor{gray!10}\multicolumn{6}{l}{\textit{\textcolor{gray}{Mixed (50\% Points + 50\% Boxes)}}\vspace{0.73mm}} \\
InternVL2 \cite{internvl2} & 26B & -- & -- & -- & 60.9 \\
Qwen2-VL \cite{qwen2vl} & 72B & -- & -- & -- & 69.1 \\
\rowcolor{blue!7.5} \textbf{\model} (Ours) & 3B & \textbf{57.2} & \textbf{64.1} & \textbf{60.6} & \underline{70.8} \\
\rowcolor{blue!7.5} \textbf{\model} (Ours) & 7B & \underline{42.3} & \underline{47.5} & \underline{44.9} & \textbf{71.4} \\
\bottomrule
\end{tabularx}
\label{tab:pixelqa}
\end{minipage}
\vspace{-4mm}
\end{table}

\vspace{-1mm}
\subsection{Q2: Pixel-Level Video Question Answering (PixelQA)}\label{sec:pixelqa}
\vspace{-1mm}

\begin{figure}
\centering
\includegraphics[width=\linewidth]{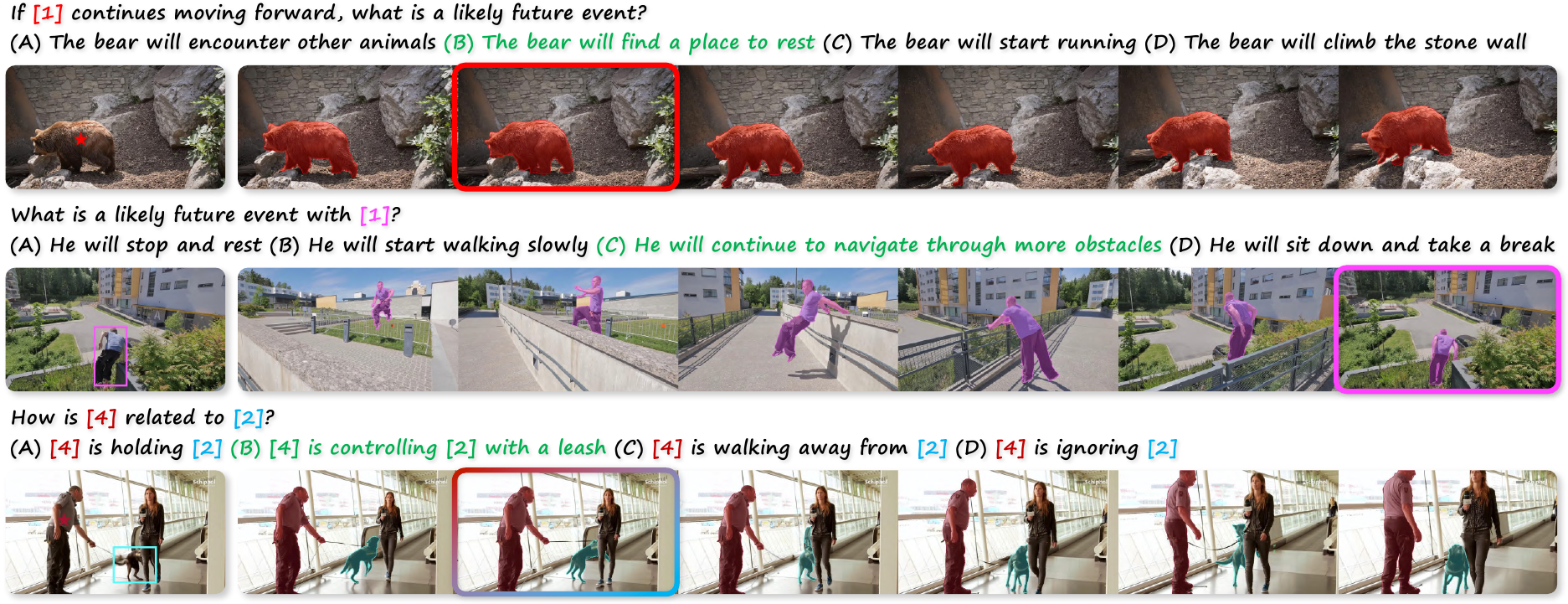}
\caption{\textbf{Visualization of the outputs from \model{} on PixelQA task}. Star marks and boxes refer to point and box prompts, respectively. The boxed frames denote where the visual prompts are applied. Given different types of visual prompts on a single frame, our method can flexibly infer the relevant object, track it across the entire video, and involve its features in reasoning.}
\label{fig:vis}
\vspace{-3mm}
\end{figure}

We design the new PixelQA task based on VideoRefer-Bench$^{\rm Q}$ \cite{videorefer}, where the original mask prompts are replaced with more challenging point or box prompts. Given these ambiguous visual cues, models are expected to correctly identify the target object according to the question and the visual prompt, then respond with \textbf{both the textual answer and the corresponding object masks}. We report the mask prediction $\mathcal{J\&F}$ and MCQ accuracy in Tab.~\ref{tab:pixelqa}. Note that none of the existing methods supports this scenario. Thus, we apply set-of-mark prompts \cite{setofmark} directly on video frames, and evaluate the QA accuracies of two strong LMMs \cite{qwen2vl,internvl2} as our baselines. Aside from point- or box-only prompts, we also explore a more flexible setting that randomly chooses different prompts for different objects. The results verify that our \textbf{\textit{memory pre-filling \& injection}} paradigm effectively enhances the model's reasoning capabilities. Visualizations of this task are shown in Fig.~\ref{fig:vis}.

\begin{table}[t]
\scriptsize
\caption{Key ablation studies with \model{}-3B on PixelQA (\textit{mixed}). See Sec.~\ref{sec:ablation} for explanations.}
\vspace{-1mm}
\begin{subtable}{0.3235\textwidth}
\caption{Task Unification}
\vspace{-1mm}
\setlength{\tabcolsep}{2.8pt}
\begin{tabularx}{\linewidth}{ccc|c|c}
\toprule
\textbf{Refer} & \textbf{Segment} & \textbf{Memory} & $\mathcal{J\&F}$ & \textbf{Acc} \\
\midrule
\cmark{} &&& -- & 64.6 \\
& \cmark{} && 47.5 & -- \\
\cmark{} & \cmark{} && 48.2 & 67.4 \\
\midrule
\rowcolor{blue!7.5} \cmark{} & \cmark{} & \cmark{} & \textbf{49.0} & \textbf{68.5} \\
\bottomrule
\end{tabularx}
\end{subtable}
\hfill
\begin{subtable}{0.3125\textwidth}
\caption{Object Memory Bank}
\vspace{-1mm}
\setlength{\tabcolsep}{3pt}
\begin{tabularx}{\linewidth}{l|c|c}
\toprule
\textbf{Referring Method} & $\mathcal{J\&F}$ & \textbf{Acc} \\
\midrule
\ding{172} \texttt{<REF>} & 46.8 & 64.5 \\
\ding{173} \texttt{<REF>}\texttt{<SEG>} & 47.8 & 64.9 \\
\ding{174} \texttt{<REF>}\texttt{<SEG>} + Pooling & 47.5 & 66.3 \\
\midrule
\rowcolor{blue!7.5} \ding{175} \textbf{Object Memory Bank} & \textbf{49.0} & \textbf{68.5} \\
\bottomrule
\end{tabularx}
\end{subtable}
\hfill
\begin{subtable}{0.3275\textwidth}
\caption{Prompt Encoder \& Mask Decoder}
\vspace{-1mm}
\setlength{\tabcolsep}{4.5pt}
\begin{tabularx}{\linewidth}{c|c|c|c}
\toprule
\textbf{Encoder} & \textbf{Decoder} & $\mathcal{J\&F}$ & \textbf{Acc} \\
\midrule
w/o Time & -- & 44.3 & 63.7 \\
\rowcolor{blue!7.5} w/ Time & -- & \textbf{49.0} & \textbf{68.5} \\
\midrule
-- & Independent & 46.1 & 66.2 \\
\rowcolor{blue!7.5} -- & Propagation & \textbf{49.0} & \textbf{68.5} \\
\bottomrule
\end{tabularx}
\end{subtable}
\label{tab:ablation}
\vspace{-3mm}
\end{table}

\subsection{Q3: Key Ablation Studies}\label{sec:ablation}

\textbf{Effect of Task Unification}\quad We study the effect of task unification in Tab.~\ref{tab:ablation}~(a). Unifying referring and segmentation capabilities into a single model and training them jointly leads to better results on both tasks (first three rows), demonstrating the \textbf{mutual reinforcement effect} of such unification. Incorporating memory pre-filling as an auxiliary task (last row) brings extra improvements.

\textbf{Effect of Object Memory Bank}\quad Tab.~\ref{tab:ablation}~(b) verifies the effectiveness of object memory bank. \ding{172} means using a single token for each referred object. \ding{173} means adding an extra segmentation token to segment it as an auxiliary task. \ding{174} further appends masked-pooled visual tokens after it. The results show that (1) both adding auxiliary segmentation task and masked-pooled features help regional understanding, and (2) decoupling them via object memory bank can further boost the performance.

\textbf{Design Space of Prompt Encoder \& Mask Decoder}\quad We compare different prompt encoder and mask decoder designs in Tab.~\ref{tab:ablation}~(c). The performance significantly drops when the temporal encoding in the prompt encoder is removed (first two rows). For the mask decoder (last two rows), we explore an alternative strategy that treats video frames independently (as batched images), which could largely accelerate inference but lead to sub-optimal accuracies. We hypothesize that this is because the LLM-generated \texttt{<SEG>} token cannot well-capture the object information in all frames, thus disentangling the segmentation and tracking capabilities to an external module is reasonable.

\vspace{-1mm}
\section{Conclusion}
\vspace{-1mm}

In this work, we proposed \textbf{\model}, a large multi-modal model that supports flexible pixel-level visual reasoning. It unifies the internal representations of referred and segmented objects through a novel \textbf{object memory bank}. We observe that by such unification, the performance of object referring and segmentation can be jointly enhanced. Extensive experiments on diverse pixel-level understanding tasks, including the \textbf{PixelQA} task, demonstrate the significance of the proposed method. We hope this work inspires future advancements in pixel-level visual understanding.

\vspace{-1mm}
\section*{Acknowledgements}
\vspace{-1mm}

This study was supported by The Hong Kong RGC Grant (15229423) and a financial support from ARC Lab, Tencent PCG (ZGG9). We also acknowledge The University Research Facility in Big Data Analytics (UBDA) at The Hong Kong Polytechnic University for providing computing resources that have contributed to the research results reported within this paper.

\clearpage

\appendix

\section*{Appendix}

In this appendix, we provide more details about the training data, model implementation, and experimental settings to complement the main paper. Additional analysis, ablation studies, visualizations, and discussions are also incorporated. Below is the table of contents.

\begin{enumerate}[label=\textbf{\Alph*.}]
\item Model
\begin{enumerate}[label=\textbf{\arabic*.}]
\item Implementation Details
\item Training Recipe
\end{enumerate}
\item Experiments
\begin{enumerate}[label=\textbf{\arabic*.}]
\item Tasks and Benchmarks
\item Evaluation Metrics
\item More Experimental Results
\item Ablation Studies
\item Qualitative Results
\end{enumerate}
\item Discussions
\begin{enumerate}[label=\textbf{\arabic*.}]
\item Limitations \& Future Work
\item Potential Societal Impacts
\end{enumerate}
\item Licenses
\end{enumerate}

\section{Model}

\subsection{Implementation Details}

We instantiate our base models with 3B and 7B versions of Qwen2.5-VL \cite{qwen2.5vl}. Both variants employ pre-trained SAM 2.1 \cite{sam2} with Hiera Base+ \cite{hiera} backbone as the mask decoder. The M$\to$L projector is initialized with the weights from the V$\to$L projector of Qwen2.5-VL. The hidden size inside the prompt encoder is 256. To reduce GPU memory and accelerate training, we randomly sample 8 frames per video, with each frame resized to 316$^{\rm 2}$\,$\sim$\,448$^{\rm 2}$ pixels (128\,$\sim$\,256 tokens per frame). The frame sampling strategies follow the specifications of each benchmark during inference. The mask decoder has a fixed resolution of 768\,$\times$\,768. For each segmentation sample, up to 5 objects are randomly selected to compute the mask prediction losses. During training, LoRA adapters \cite{lora} with \texttt{rank=128} and \texttt{alpha=256} are applied to all \texttt{QKVO} layers in the visual encoder and LLM. The input sequences are restricted to 4K tokens. We train the model with 8 RTX A6000 Ada (48G) GPUs, with a global batch size of 256 for stages 1 and 2, and 32 for stage 3. In the first two stages, the learning rates are set to 1e-3. In the last stage, it is set to 5e-6 for the mask decoder and 2e-5 for all the other parameters, respectively. A linear warmup in the first 3\% steps followed by cosine decay is adopted in all stages. The configurations of datasets are introduced in the following section.

\subsection{Training Recipe}

The detailed distribution of training datasets for \model{} is shown in Tab.~\ref{tab:training}. Within the three-stage training recipe, we first pre-train the sparse prompt encoder using short caption samples from Inst-IT \cite{instit} and VideoRefer \cite{videorefer}. For each sample, we randomly select a point inside the ground truth mask (50\%) or generate an augmented box from it (50\%). This stage aims to enable the model with simple visual prompt comprehension and regional captioning capabilities on images and videos. In the second stage, we align the LLM and mask decoder using referring object segmentation datasets \cite{refcoco,refcocog,urvos}. We use short caption/query samples for the first two stages to focus on alignment rather than knowledge learning. For the last stage, we collect a large-scale, high-quality corpus called \texttt{UniPixel-SFT-1M}\footnote{\,\url{https://huggingface.co/datasets/PolyU-ChenLab/UniPixel-SFT-1M}} to jointly train the model on diverse pixel-level tasks. The original annotations have been rewritten using task-specific templates with instructions. General image and video understanding datasets \cite{llava1.5,videogptplus} are also incorporated, resulting around 2M samples in total.

\begin{table}
\scriptsize
\setlength{\tabcolsep}{3.625pt}
\centering
\caption{The distribution of training datasets for \model{}. We use different background colors to denote \colorbox{red!6.5}{object referring}, \colorbox{blue!3.5}{object segmentation}, \colorbox{orange!8.5}{regional understanding}, \colorbox{green!6.5}{memory pre-filling}, and \colorbox{yellow!10}{general image \& video understanding} data, respectively.}
\vspace{1.2mm}
\begin{tabularx}{\linewidth}{c|p{3.45cm}<{\raggedright}|cccccc|cc|rrr}
\toprule
\multirow{2.6}{*}{\textbf{Stage}} & \multirow{2.6}{*}{\textbf{Dataset}} & \multicolumn{6}{c|}{\textbf{Inputs}} & \multicolumn{2}{c|}{\textbf{Outputs}} & \multirow{2.6}{*}{\textbf{\#Samples}} & \multirow{2.6}{*}{\textbf{\#Repeat}} & \multirow{2.6}{*}{\textbf{Ratio}} \\
\cmidrule{3-8} \cmidrule{9-10}
&& Text & Image & Video & Point & Box & Mask & Text & Mask \\
\midrule
\rowcolor{red!6.5}\cellcolor{white}\multirow{2}{*}{1} & Inst-IT-Image-Short-Caption \cite{instit} & \cmark & \cmark && \cmark & \cmark && \cmark && 351K & 1 & 41.2\% \\
& \cellcolor{red!6.5}VideoRefer-Short-Caption \cite{videorefer} & \cellcolor{red!6.5}\cmark & \cellcolor{red!6.5} & \cellcolor{red!6.5}\cmark & \cellcolor{red!6.5}\cmark & \cellcolor{red!6.5}\cmark & \cellcolor{red!6.5} & \cellcolor{red!6.5}\cmark & \cellcolor{red!6.5} & \cellcolor{red!6.5}500K & \cellcolor{red!6.5}1 & \cellcolor{red!6.5}58.8\% \\
\midrule
\rowcolor{blue!3.5}\cellcolor{white} & RefCOCO \cite{refcoco} & \cmark & \cmark &&&&& \cmark & \cmark & 17K & 5 & 20.8\% \\
\rowcolor{blue!3.5}\cellcolor{white} & RefCOCO+ \cite{refcoco} & \cmark & \cmark &&&&& \cmark & \cmark & 17K & 5 & 20.8\% \\
\rowcolor{blue!3.5}\cellcolor{white}2 & RefCOCOg \cite{refcocog} & \cmark & \cmark &&&&& \cmark & \cmark & 22K & 5 & 26.8\% \\
\rowcolor{blue!3.5}\cellcolor{white} & RefClef \cite{refcoco} & \cmark & \cmark &&&&& \cmark & \cmark & 18K & 5 & 22.0\% \\
\rowcolor{blue!3.5}\cellcolor{white} & Ref-YouTube-VOS \cite{urvos} & \cmark && \cmark &&&& \cmark & \cmark & 13K & 3 & 9.5\% \\
\midrule
\rowcolor{orange!8.5}\cellcolor{white}\multirow{28}{*}{3} & Osprey-Conversation \cite{osprey} & \cmark & \cmark &&&& \cmark & \cmark && 1.4K & 5 & 0.1\% \\
\rowcolor{orange!8.5}\cellcolor{white} & Osprey-Detail-Description \cite{osprey} & \cmark & \cmark &&&& \cmark & \cmark && 29K & 5 & 2.5\% \\
\rowcolor{orange!8.5}\cellcolor{white} & Osprey-Pos-Neg \cite{osprey} & \cmark & \cmark &&&& \cmark & \cmark && 20K & 5 & 1.7\% \\
\rowcolor{orange!8.5}\cellcolor{white} & VideoRefer-Detailed-Caption \cite{videorefer} & \cmark && \cmark &&& \cmark & \cmark && 120K & 5 & 10.1\% \\
\rowcolor{orange!8.5}\cellcolor{white} & VideoRefer-QA \cite{videorefer} & \cmark && \cmark &&& \cmark & \cmark && 69K & 5 & 5.8\% \\
\rowcolor{orange!8.5}\cellcolor{white} & Inst-IT-Video-QA \cite{instit} & \cmark && \cmark &&& \cmark & \cmark && 159K & 5 & 13.4\% \\
\rowcolor{green!6.5}\cellcolor{white} & VideoRefer-QA-Memory \cite{videorefer} & \cmark && \cmark & \cmark & \cmark && \cmark & \cmark & 69K & 3 & 3.5\% \\
\rowcolor{green!6.5}\cellcolor{white} & Inst-IT-QA-Memory \cite{instit} & \cmark && \cmark & \cmark & \cmark && \cmark & \cmark & 158K & 3 & 8.0\% \\
\rowcolor{blue!3.5}\cellcolor{white} & RefCOCO \cite{refcoco} & \cmark & \cmark &&&&& \cmark & \cmark & 17K & 10 & 2.9\% \\
\rowcolor{blue!3.5}\cellcolor{white} & RefCOCO+ \cite{refcoco} & \cmark & \cmark &&&&& \cmark & \cmark & 17K & 10 & 2.9\% \\
\rowcolor{blue!3.5}\cellcolor{white} & RefCOCOg \cite{refcocog} & \cmark & \cmark &&&&& \cmark & \cmark & 22K & 10 & 3.7\% \\
\rowcolor{blue!3.5}\cellcolor{white} & RefClef \cite{refcoco} & \cmark & \cmark &&&&& \cmark & \cmark & 18K & 10 & 3.0\% \\
\rowcolor{blue!3.5}\cellcolor{white} & ReasonSeg \cite{lisa} & \cmark & \cmark &&&&& \cmark & \cmark & 1.6K & 10 & 0.3\% \\
& \cellcolor{blue!3.5}ADE20K \cite{ade20k} & \cellcolor{blue!3.5}\cmark & \cellcolor{blue!3.5}\cmark & \cellcolor{blue!3.5} & \cellcolor{blue!3.5} & \cellcolor{blue!3.5} & \cellcolor{blue!3.5} & \cellcolor{blue!3.5}\cmark & \cellcolor{blue!3.5}\cmark & \cellcolor{blue!3.5}20K & \cellcolor{blue!3.5}3 & \cellcolor{blue!3.5}1.0\% \\
& \cellcolor{blue!3.5}COCOStuff \cite{cocostuff} & \cellcolor{blue!3.5}\cmark & \cellcolor{blue!3.5}\cmark & \cellcolor{blue!3.5} & \cellcolor{blue!3.5} & \cellcolor{blue!3.5} & \cellcolor{blue!3.5} & \cellcolor{blue!3.5}\cmark & \cellcolor{blue!3.5}\cmark & \cellcolor{blue!3.5}118K & \cellcolor{blue!3.5}3 & \cellcolor{blue!3.5}6.0\% \\
\rowcolor{blue!3.5}\cellcolor{white} & Mapillary\,Vistas \cite{mapillary} & \cmark & \cmark &&&&& \cmark & \cmark & 18K & 3 & 0.9\% \\
\rowcolor{blue!3.5}\cellcolor{white} & PACO-LVIS \cite{paco} & \cmark & \cmark &&&&& \cmark & \cmark & 46K & 3 & 2.3\% \\
\rowcolor{blue!3.5}\cellcolor{white} & PASCAL-Part \cite{pascalpart} & \cmark & \cmark &&&&& \cmark & \cmark & 4.4K & 3 & 0.2\% \\
\rowcolor{blue!3.5}\cellcolor{white} & Ref-YouTube-VOS \cite{urvos} & \cmark && \cmark &&&& \cmark & \cmark & 13K & 5 & 1.1\% \\
\rowcolor{blue!3.5}\cellcolor{white} & Ref-DAVIS17 \cite{davis2017} & \cmark && \cmark &&&& \cmark & \cmark & 0.6K & 10 & 0.1\% \\
\rowcolor{blue!3.5}\cellcolor{white} & Ref-SAV \cite{sa2va} & \cmark && \cmark &&&& \cmark & \cmark & 56K & 3 & 2.8\% \\
\rowcolor{blue!3.5}\cellcolor{white} & MeViS \cite{mevis} & \cmark && \cmark &&&& \cmark & \cmark & 23K & 5 & 1.9\% \\
\rowcolor{blue!3.5}\cellcolor{white} & LV-VIS \cite{lvvis} & \cmark && \cmark &&&& \cmark & \cmark & 11K & 3 & 0.6\% \\
\rowcolor{blue!3.5}\cellcolor{white} & ViCaS \cite{vicas} & \cmark && \cmark &&&& \cmark & \cmark & 41K & 3 & 2.1\% \\
\rowcolor{blue!3.5}\cellcolor{white} & ReVOS \cite{visa} & \cmark && \cmark &&&& \cmark & \cmark & 29K & 5 & 2.5\% \\
\rowcolor{blue!3.5}\cellcolor{white} & GroundMoRe \cite{groundmore} & \cmark && \cmark &&&& \cmark & \cmark & 5.6K & 3 & 0.3\% \\
\rowcolor{yellow!10}\cellcolor{white} & LLaVA-1.5-Mix-665K \cite{llava1.5} & \cmark & \cmark &&&&& \cmark && 647K & 1 & 10.9\% \\
\rowcolor{yellow!10}\cellcolor{white} & VideoGPT+ Instruct \cite{videogptplus} & \cmark && \cmark &&&& \cmark && 573K & 1 & 9.7\% \\
\bottomrule
\end{tabularx}
\label{tab:training}
\end{table}

\section{Experiments}

\subsection{Tasks and Benchmarks}

Our method is extensively evaluated across 9 fine-grained image/video understanding tasks. The benchmark(s) used for each task are listed as follows:

\begin{enumerate}[left=0pt,itemsep=0pt]
\fontsize{8.5pt}{8.5pt}\selectfont
\item \textbf{Reasoning Video Object Segmentation:} ReVOS \cite{visa}
\item \textbf{Referring Video Object Segmentation:} MeViS \cite{mevis}, Ref-YouTube-VOS \cite{urvos}, Ref-DAVIS17 \cite{davis2017}, Ref-SAV \cite{sa2va}
\item \textbf{Motion-Grounded Video Reasoning:} GroundMoRe \cite{groundmore}
\item \textbf{Referring Expression Segmentation:} RefCOCO \cite{refcoco}, RefCOCO+ \cite{refcoco}, RefCOCOg \cite{refcocog}
\item \textbf{Reasoning Segmentation:} ReasonSeg \cite{lisa}
\item \textbf{Referring Expression Comprehension:} RefCOCO \cite{refcoco}, RefCOCO+ \cite{refcoco}, RefCOCOg \cite{refcocog}
\item \textbf{Referred Video Description:} VideoRefer-Bench$^{\rm D}$ \cite{videorefer}
\item \textbf{Referred Video Question Answering:} VideoRefer-Bench$^{\rm Q}$ \cite{videorefer}
\item \textbf{Flexible Pixel-Level Understanding:} PixelQA (Ours)
\end{enumerate}

\subsection{Evaluation Metrics}

For video segmentation tasks, we adopt $\mathcal{J\&F}$ as the main metric to jointly consider region similarity $\mathcal{J}$ and contour accuracy $\mathcal{F}$. Image segmentation is evaluated using cIoU (the cumulative intersection over the cumulative union) and gIoU (the average of all per-image IoUs) following existing work. For referred video description and question answering tasks, we follow the official evaluation protocols to report GPT-4o \cite{gpt4o} scores and MCQ accuracy, respectively. For referring expression comprehension, we leverage mean accuracies, where a predicted bounding box is considered correct when it has the intersection over union (IoU) with the ground truth no less than 0.5.

\subsection{More Experimental Results}

\textbf{General Video Question Answering}\quad We also evaluate \model{} on MVBench \cite{mvbench} to compare its general video understanding capabilities with existing methods. The results are illustrated in Tab.~\ref{tab:mvbench}. Note that our method is the only one in the table that supports referring and segmentation. By jointly training on holistic-level and pixel-level data, \model{} can effectively balance the capabilities under both scenarios, demonstrated by the strong performance compared with holistic-level models.

\begin{table}
\scriptsize
\centering
\setlength{\tabcolsep}{1.565pt}
\fontsize{6.9pt}{7.5pt}\selectfont
\caption{Performance comparison on general video question answering (VideoQA) on MVBench \cite{mvbench}. Note that \model{} is the only model supporting pixel-level referring \& segmentation.}
\vspace{2mm}
\begin{tabularx}{\linewidth}{l|c|cccccccccccccccccccc|c}
\toprule
\textbf{Model} & \textbf{Size} & \textbf{AS} & \textbf{AP} & \textbf{AA} & \textbf{FA} & \textbf{UA} & \textbf{OE} & \textbf{OI} & \textbf{OS} & \textbf{MD} & \textbf{AL} & \textbf{ST} & \textbf{AC} & \textbf{MC} & \textbf{MA} & \textbf{SC} & \textbf{FP} & \textbf{CO} & \textbf{EN} & \textbf{ER} & \textbf{CI} & \textbf{Avg.} \\
\midrule
GPT-4V \cite{gpt4v} & -- & 55.5 & \underline{63.5} & 72.0 & 46.5 & 73.5 & 18.5 & 59.0 & 29.5 & 12.0 & 40.5 & 83.5 & 39.0 & 12.0 & 22.5 & 45.0 & 47.5 & 52.0 & 31.0 & \textbf{59.0} & 11.0 & 43.5 \\
\midrule
Video-ChatGPT \cite{videochatgpt} & 7B & 23.5 & 26.0 & 62.0 & 22.5 & 26.5 & 54.0 & 28.0 & 40.0 & 23.0 & 20.0 & 31.0 & 30.5 & 25.5 & 39.5 & 48.5 & 29.0 & 33.0 & 29.5 & 26.0 & 35.5 & 32.7 \\
Video-LLaMA \cite{videollama} & 7B & 27.5 & 25.5 & 51.0 & 29.0 & 39.0 & 48.0 & 40.5 & 38.0 & 22.5 & 22.5 & 43.0 & 34.0 & 22.5 & 32.5 & 45.5 & 32.5 & 40.0 & 30.0 & 21.0 & 37.0 & 34.1 \\
VideoChat \cite{videochat} & 7B & 33.5 & 26.5 & 56.0 & 33.5 & 40.5 & 53.0 & 40.5 & 30.0 & 25.5 & 27.0 & 48.5 & 35.0 & 20.5 & 42.5 & 46.0 & 26.5 & 41.0 & 23.5 & 23.5 & 36.0 & 35.5 \\
Video-LLaVA \cite{videollava} & 7B & 46.0 & 42.5 & 56.5 & 39.0 & 53.5 & 53.0 & 48.0 & \underline{41.0} & 29.0 & 31.5 & 82.5 & 45.0 & 26.0 & 53.0 & 41.5 & 33.5 & 41.5 & 27.5 & 38.5 & 31.5 & 43.0 \\
TimeChat \cite{timechat} & 7B & 40.5 & 36.0 & 61.0 & 32.5 & 53.0 & 53.5 & 41.5 & 29.0 & 19.5 & 26.5 & 66.5 & 34.0 & 20.0 & 43.5 & 42.0 & 36.5 & 36.0 & 29.0 & 35.0 & 35.0 & 38.5 \\
PLLaVA \cite{pllava} & 7B & 58.0 & 49.0 & 55.5 & 41.0 & 61.0 & 56.0 & 61.0 & 36.0 & 23.5 & 26.0 & 82.0 & 39.5 & 42.0 & 52.0 & 45.0 & 42.0 & 53.5 & 30.5 & 48.0 & 31.0 & 46.6 \\
ST-LLM \cite{stllm} & 7B & 66.0 & 53.5 & \textbf{84.0} & 44.0 & 58.5 & 80.5 & 73.5 & 38.5 & 42.5 & 31.0 & 86.5 & 36.5 & 56.5 & 78.5 & 43.0 & 44.5 & 46.5 & 34.5 & 41.5 & 58.5 & 54.9 \\
VideoGPT+ \cite{videogptplus} & 4B & 69.0 & 60.0 & 83.0 & \underline{48.5} & 66.5 & 85.5 & \textbf{75.5} & 36.0 & 44.0 & 34.0 & \underline{89.5} & 39.5 & 71.0 & 90.5 & 45.0 & \underline{53.0} & 50.0 & 29.5 & 44.0 & 60.0 & 58.7 \\
VideoChat2 \cite{mvbench} & 7B & \textbf{75.5} & 58.0 & \underline{83.5} & \textbf{50.5} & 60.5 & \underline{87.5} & \underline{74.5} & \textbf{45.0} & 47.5 & \textbf{44.0} & 82.5 & 37.0 & 64.5 & 87.5 & \underline{51.0} & \textbf{66.5} & 47.0 & \textbf{35.0} & 37.0 & \underline{72.5} & 60.4 \\
\midrule
\rowcolor{blue!7.5} \textbf{\model} (Ours) & 3B & 69.5 & 62.5 & 83.0 & \underline{48.5} & \underline{76.5} & 86.5 & 66.5 & 38.0 & \underline{49.0} & 40.5 & 87.0 & \textbf{49.0} & \textbf{74.0} & \textbf{95.0} & 49.0 & 45.0 & \underline{63.5} & 34.5 & \underline{58.0} & \textbf{73.5} & \underline{62.5} \\
\rowcolor{blue!7.5} \textbf{\model} (Ours) & 7B & \underline{71.0} & \textbf{68.0} & \textbf{84.0} & 45.0 & \textbf{78.0} & \textbf{91.5} & 66.5 & 35.5 & \textbf{57.5} & \underline{43.0} & \textbf{91.5} & \underline{47.0} & \underline{73.5} & \underline{92.5} & \textbf{58.0} & \underline{53.0} & \textbf{74.0} & \textbf{37.5} & 49.0 & 69.0 & \textbf{64.3} \\
\bottomrule
\end{tabularx}
\label{tab:mvbench}
\vspace{-3mm}
\end{table}

\subsection{Ablation Studies}

\textbf{Effect of Multi-stage Training}\quad We investigate the effectiveness of multi-stage training in Tab.~\ref{tab:ablation_stage}. As shown in the first line, directly training the model using large-scale data only leads to sub-optimal performance, due to the unaligned representations among prompt encoder, LLM, and mask decoder. We observe that pre-training either the sparse prompt encoder or the L$\to$M projector (the second and third lines) brings performance gains on both tasks (referring and segmentation). We hypothesize that this is because pre-aligning either of them can alleviate the burden of joint-task learning in stage 3. The last row verifies that the performance can be further boosted by pre-aligning both of them.

\begin{table}
\scriptsize
\centering
\caption{Effectiveness justification of multi-stage training. The best and second-best results are marked \textbf{bold} and \underline{underlined}, respectively. The three-stage recipe leads to optimal performance.}
\vspace{2mm}
\begin{tabularx}{0.905\linewidth}{ccc|ccc|ccc|cc}
\toprule
\multirow{2.6}{*}{\textbf{Stage 1}} & \multirow{2.6}{*}{\textbf{Stage 2}} & \multirow{2.6}{*}{\textbf{Stage 3}} & \multicolumn{3}{c|}{\textbf{ReVOS}} & \multicolumn{3}{c|}{\textbf{MeViS} (\texttt{val$^{\texttt{u}}$})} & \multicolumn{2}{c}{\textbf{VideoRefer-Bench$^{\rm Q}$}} \\
\cmidrule{4-6} \cmidrule{7-9} \cmidrule{10-11}
&&& $\mathcal{J}$ & $\mathcal{F}$ & $\mathcal{J\&F}$ & $\mathcal{J}$ & $\mathcal{F}$ & $\mathcal{J\&F}$ & Single-Frame & Multi-Frame \\
\midrule
&& \cmark & 58.3 & \underline{63.6} & 61.0 & 54.8 & 61.9 & 58.4 & 71.1 & 71.5 \\
\cmark && \cmark & 59.0 & 63.4 & 61.2 & 55.2 & 62.1 & 58.7 & \underline{71.8} & \underline{72.3} \\
& \cmark & \cmark & \underline{59.6} & 63.5 & \underline{61.6} & \underline{55.7} & \underline{62.5} & \underline{59.1} & 71.2 & 71.6 \\
\midrule
\rowcolor{blue!7.5} \cmark & \cmark & \cmark & \textbf{59.7} & \textbf{64.4} & \textbf{62.1} & \textbf{56.1} & \textbf{63.2} & \textbf{59.7} & \textbf{72.2} & \textbf{72.8} \\
\bottomrule
\end{tabularx}
\label{tab:ablation_stage}
\vspace{-3mm}
\end{table}

\textbf{Number of Hidden Tokens for Mask Decoder}\quad There is a huge gap between the feature dimensions of the LLM and the mask decoder, thus splitting the \texttt{<SEG>} token into more hidden tokens can better preserve the object information from the LLM. We ablate this mechanism in Tab.~\ref{tab:ablation_seg_token}. According to the results, using only 1 hidden token cannot fully preserve the object information, as the mask prediction performance is sub-optimal. However, we also observe that using more than 2 hidden tokens (\eg, 4 or 8) only brings negligible performance gain. Therefore, we choose 2 hidden tokens per object in our final model.

\textbf{Training Strategy for the M$\to$L projector}\quad The M$\to$L projector aims to project the masked-pooled object-centric features to the LLM's embedding space. Since the object features originate from the visual encoder, it is possible to re-use the pre-trained weights of the original V$\to$L projector in Qwen2.5-VL. Its effects are studied in Tab.~\ref{tab:ablation_msk_proj}. We investigated two strategies: 1) re-using the weights and 2) adding an extra pre-training stage for better alignment. The comparison shows that directly re-using weights without extra pre-training can achieve the best results.

\textbf{Combination of Training Data}\quad Tab.~\ref{tab:ablation_data} studies the effect of the combination of multi-task co-training data in stage 3. Compared with training only on the regional or segmentation data, leveraging both of them leads to considerable performance on both tasks. Incorporating memory pre-filling data (requiring both referring and segmentation) can further boost the performance. We also mix some general holistic-level video understanding data to preserve the original capabilities of the pre-trained model, while it slightly affects the performance on pixel-level tasks.

\subsection{Qualitative Results}

\begin{figure}[t]
\centering
\includegraphics[width=\linewidth]{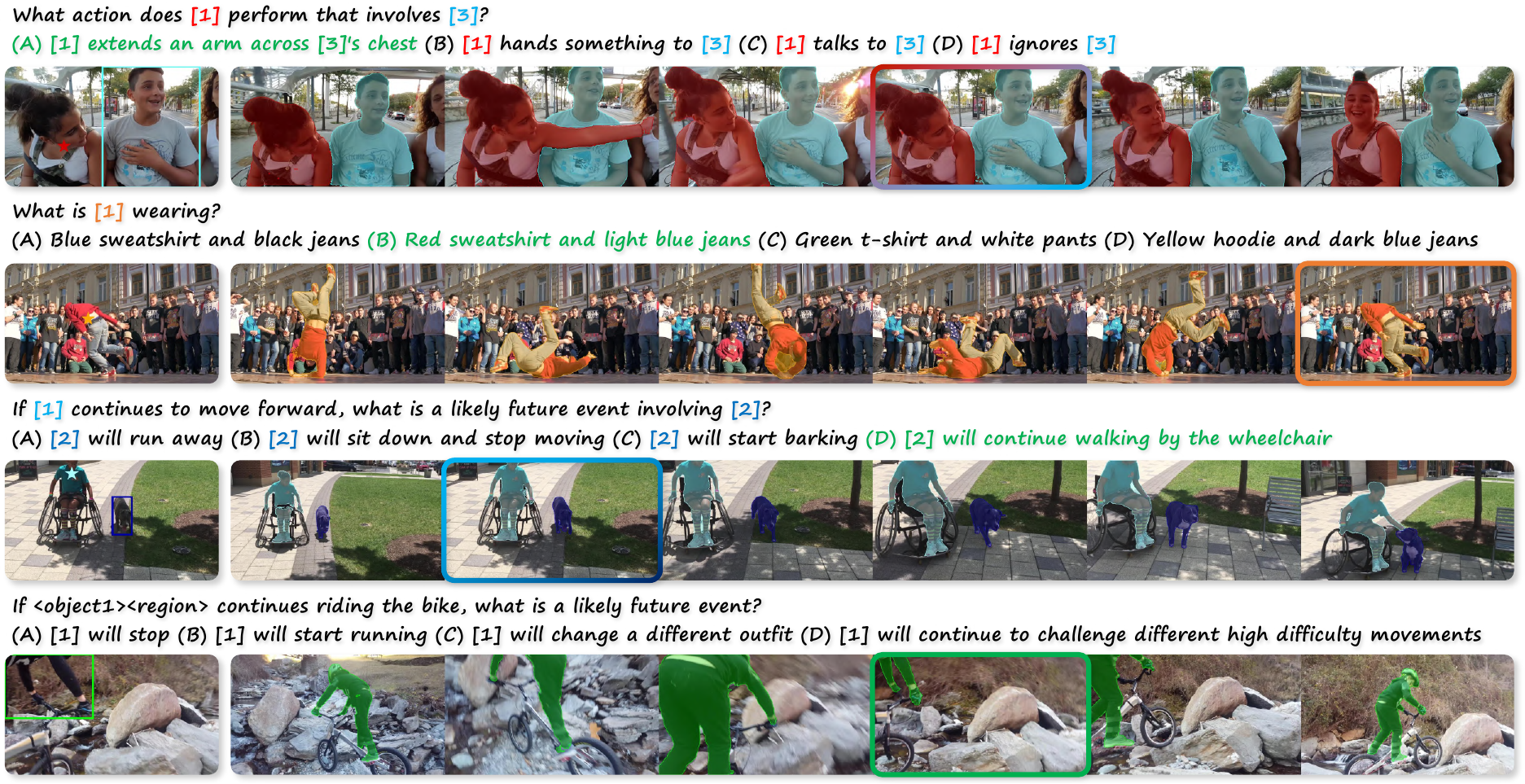}
\caption{Visualization of the predictions from \model{} on PixelQA.}
\label{fig:vis_pixel}
\end{figure}

\begin{figure}[t]
\centering
\includegraphics[width=\linewidth]{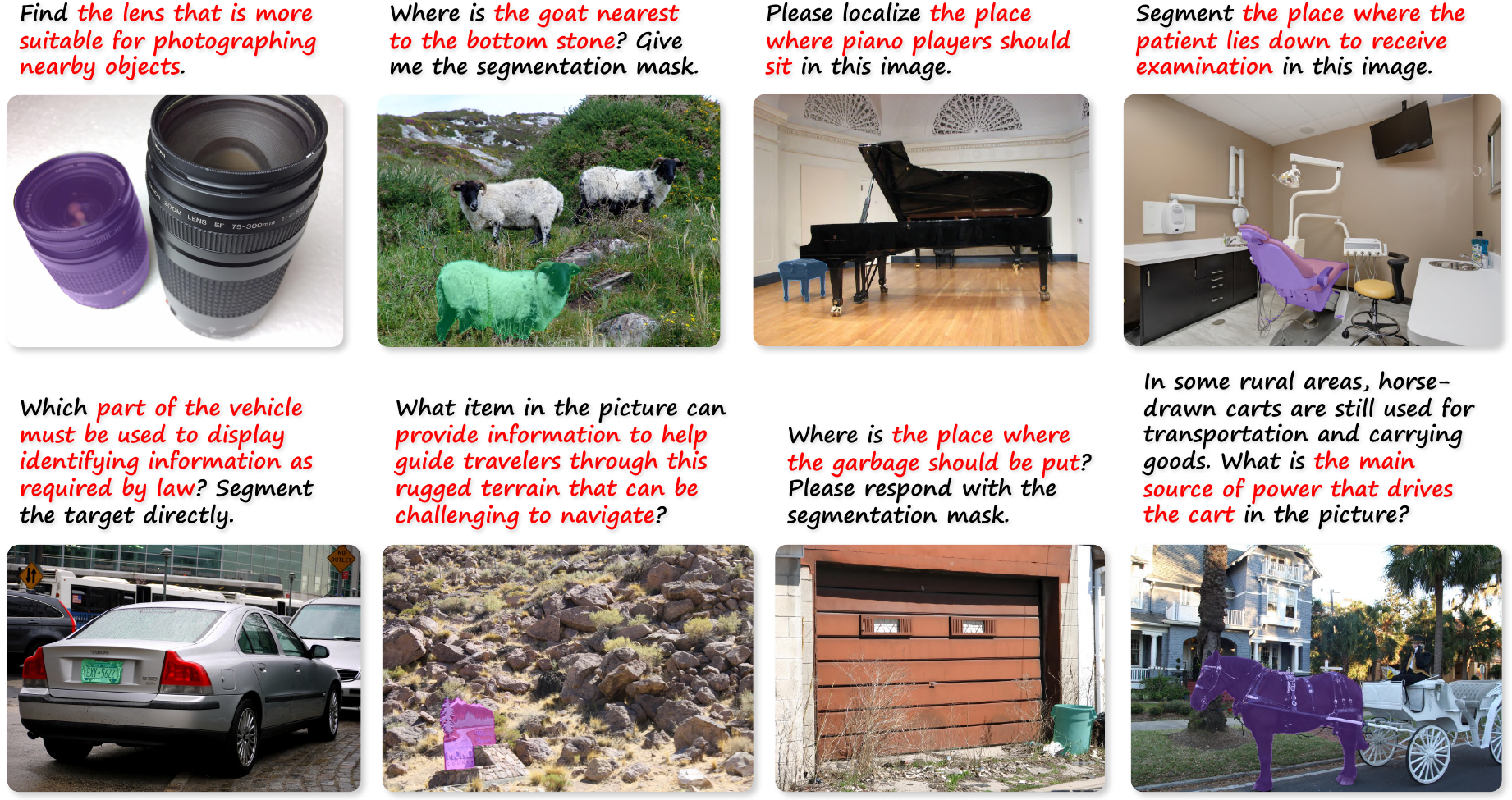}
\caption{Visualization of the predictions from \model{} on ReasonSeg \cite{lisa}.}
\label{fig:vis_reason}
\end{figure}

Fig.~\ref{fig:vis_pixel}\,$\sim$\,\ref{fig:vis_reason} present more visualizations of outputs from \model{} on different pixel-level understanding tasks. Our method can effectively handle flexible visual prompts \cite{videorefer}, implicit queries \cite{lisa,visa}, long queries \cite{sa2va}, and motion-grounded questions \cite{groundmore}.

\begin{table}
\scriptsize
\begin{minipage}{0.48\textwidth}
\caption{Ablation study on the number of hidden tokens for each \texttt{<SEG>}. Performance gains are negligible with more than 2 tokens/object.}
\vspace{2mm}
\setlength{\tabcolsep}{5.535pt}
\begin{tabularx}{\linewidth}{c|ccc|ccc}
\toprule
\multirow{2.6}{*}{\textbf{\#Tokens}} & \multicolumn{3}{c|}{\textbf{ReVOS}} & \multicolumn{3}{c}{\textbf{MeViS} (\texttt{val$^{\texttt{u}}$})} \\
\cmidrule{2-4} \cmidrule{5-7}
& $\mathcal{J}$ & $\mathcal{F}$ & $\mathcal{J\&F}$ & $\mathcal{J}$ & $\mathcal{F}$ & $\mathcal{J\&F}$ \\
\midrule
1 & 59.6 & 63.5 & 61.6 & 55.8 & 62.5 & 59.2 \\
\rowcolor{blue!7.5} 2 & \textbf{59.7} & \textbf{64.4} & \textbf{62.1} & 56.1 & \textbf{63.2} & \underline{59.7} \\
4 & \underline{59.8} & 63.9 & \underline{61.9} & \textbf{56.8} & \underline{63.1} & \textbf{59.9} \\
8 & 59.5 & \underline{64.0} & 61.8 & \underline{56.4} & \underline{62.8} & \underline{59.6} \\
\bottomrule
\end{tabularx}
\label{tab:ablation_seg_token}
\end{minipage}
\hfill
\begin{minipage}{0.48\textwidth}
\caption{Ablation study on M$\to$L projector. \underline{Init} and \underline{PT} denote weight initialization from V$\to$L projector and extra pre-training, respectively.}
\vspace{2mm}
\setlength{\tabcolsep}{6.5pt}
\begin{tabularx}{\linewidth}{cc|cc|c}
\toprule
\multirow{2.6}{*}{\textbf{Init}} & \multirow{2.6}{*}{\textbf{PT}} & \multicolumn{2}{c|}{\textbf{VideoRefer-Bench$^{\rm Q}$}} & \textbf{PixelQA} \\
\cmidrule{3-4} \cmidrule{5-5}
&& Single-Frame & Multi-Frame & Mixed Acc \\
\midrule
&& 71.4 & 71.9 & 67.7 \\
& \cmark & 71.5 & 71.7 & 67.4 \\
\rowcolor{blue!7.5} \cmark && \textbf{72.4} & \underline{72.6} & \underline{68.2} \\
\cmark & \cmark & \underline{72.2} & \textbf{72.8} & \textbf{68.5} \\
\bottomrule
\end{tabularx}
\label{tab:ablation_msk_proj}
\end{minipage}
\vspace{-3mm}
\end{table}

\begin{table}
\scriptsize
\centering
\setlength{\tabcolsep}{4.9pt}
\caption{Ablation study on training data used in stage 3. The best and second-best results are marked \textbf{bold} and \underline{underlined}, respectively. Gradually adding more pixel-level data brings performance gains.}
\vspace{2mm}
\begin{tabularx}{\linewidth}{cccc|ccc|ccc|cc}
\toprule
\multirow{2.6}{*}{\textbf{Regional}} & \multirow{2.6}{*}{\textbf{Segmentation}} & \multirow{2.6}{*}{\textbf{Memory}} & \multirow{2.6}{*}{\textbf{General}} & \multicolumn{3}{c|}{\textbf{ReVOS}} & \multicolumn{3}{c|}{\textbf{MeViS} (\texttt{val$^{\texttt{u}}$})} & \multicolumn{2}{c}{\textbf{VideoRefer-Bench$^{\rm Q}$}} \\
\cmidrule{5-7} \cmidrule{8-10} \cmidrule{11-12}
&&&& $\mathcal{J}$ & $\mathcal{F}$ & $\mathcal{J\&F}$ & $\mathcal{J}$ & $\mathcal{F}$ & $\mathcal{J\&F}$ & Single-Frame & Multi-Frame \\
\midrule
\cmark &&&& -- & -- & -- & -- & -- & -- & 72.1 & 72.0 \\
& \cmark &&& 58.9 & 63.8 & 61.4 & 56.0 & \underline{63.2} & 59.6 & -- & -- \\
\cmark & \cmark &&& 59.2 & 63.7 & 61.5 & 55.8 & 63.1 & 59.5 & \underline{72.3} & \underline{72.6} \\
\cmark & \cmark & \cmark && \underline{59.6} & \textbf{64.5} & \textbf{62.1} & \textbf{56.3} & \textbf{63.5} & \textbf{59.9} & \textbf{72.4} & 72.5 \\
\midrule
\rowcolor{blue!7.5} \cmark & \cmark & \cmark & \cmark & \textbf{59.7} & \underline{64.4} & \textbf{62.1} & \underline{56.1} & \underline{63.2} & \underline{59.7} & 72.2 & \textbf{72.8} \\
\bottomrule
\end{tabularx}
\label{tab:ablation_data}
\vspace{-4mm}
\end{table}

\vspace{-1mm}
\section{Discussion}

\subsection{Limitations \& Future Work}

Due to the limited computing resources, we did not further scale up the training data to incorporate more pixel-level tasks such as grounded caption generation (GCG) on images \cite{glamm} or videos \cite{videoglamm}, which are interesting scenarios and their data may bring more performance gains. Besides, the mask decoder currently predicts the first mask on the first frame and propagates it to the following frames, while it potentially supports predicting on the best frame (defined as the frame with the best view of the target) and propagates it to both sides of the video. We will focus in our future work to explore more pixel-level understanding tasks and more flexible mechanisms for the mask decoder.

\subsection{Potential Societal Impacts}

This work introduces a new framework for pixel-level visual-language understanding, which could potentially be used in education, surveillance, and healthcare industries, where flexible interactions with the users and fine-grained understanding of images \& videos are required. In other scenarios requiring multi-modal assistants, our method can also serve as a more advanced alternative. To the best of our knowledge, there are no potential negative societal impacts to declare.

\vspace{-1mm}
\section{Licenses}

Our model is built based on the pre-trained Qwen2.5-VL \cite{qwen2.5vl} and SAM 2.1 \cite{sam2} models. They are both licensed under the Apache License 2.0 (\url{https://www.apache.org/licenses/LICENSE-2.0}).

\begin{figure}[t]
\centering
\includegraphics[width=\linewidth]{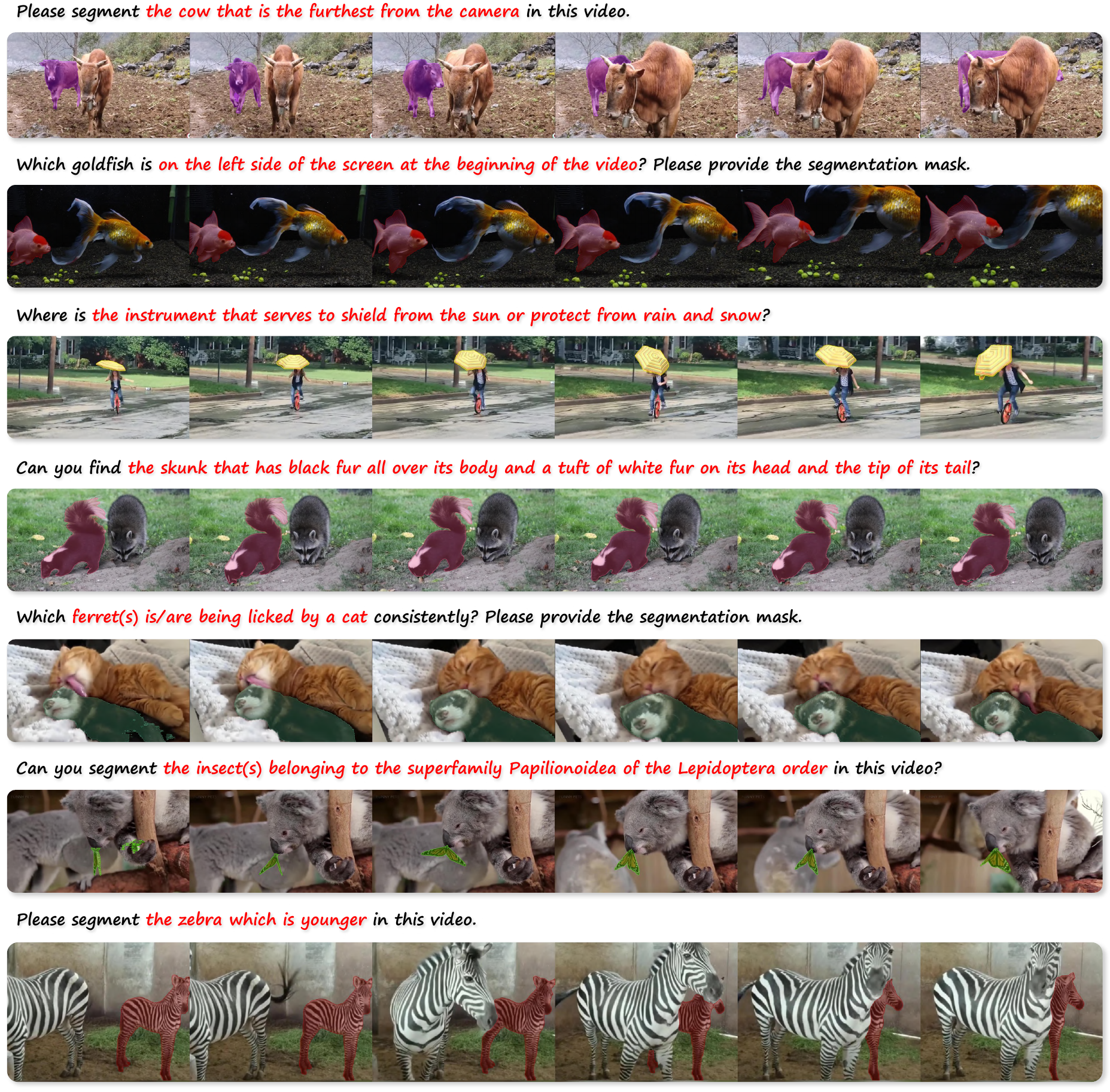}
\caption{Visualization of the predictions from \model{} on ReVOS \cite{visa}.}
\label{fig:vis_revos}
\end{figure}

\begin{figure}[t]
\centering
\includegraphics[width=\linewidth]{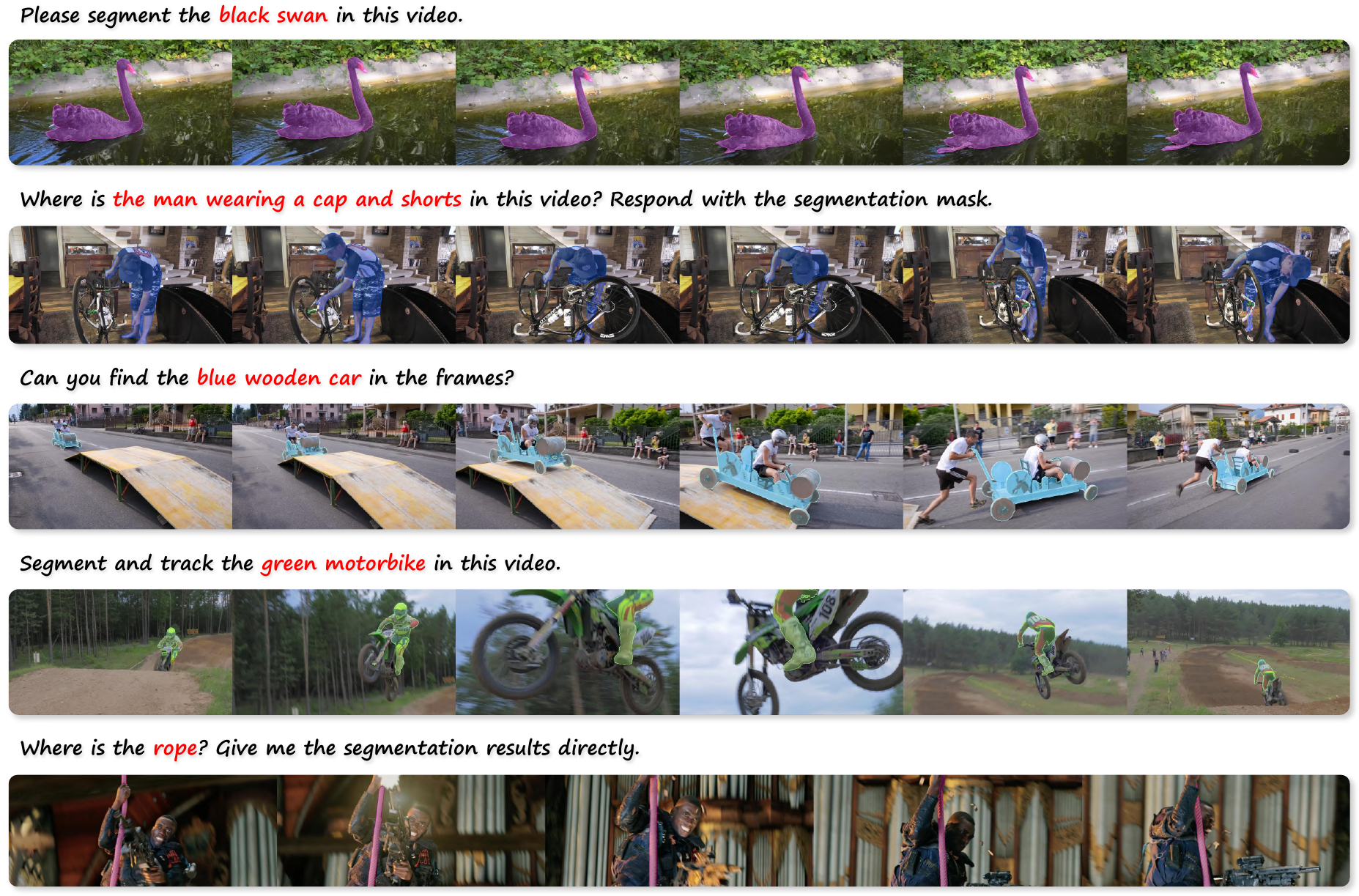}
\caption{Visualization of the predictions from \model{} on Ref-DAVIS17 \cite{davis2017}.}
\label{fig:vis_rvos}
\end{figure}

\begin{figure}[t]
\centering
\includegraphics[width=\linewidth]{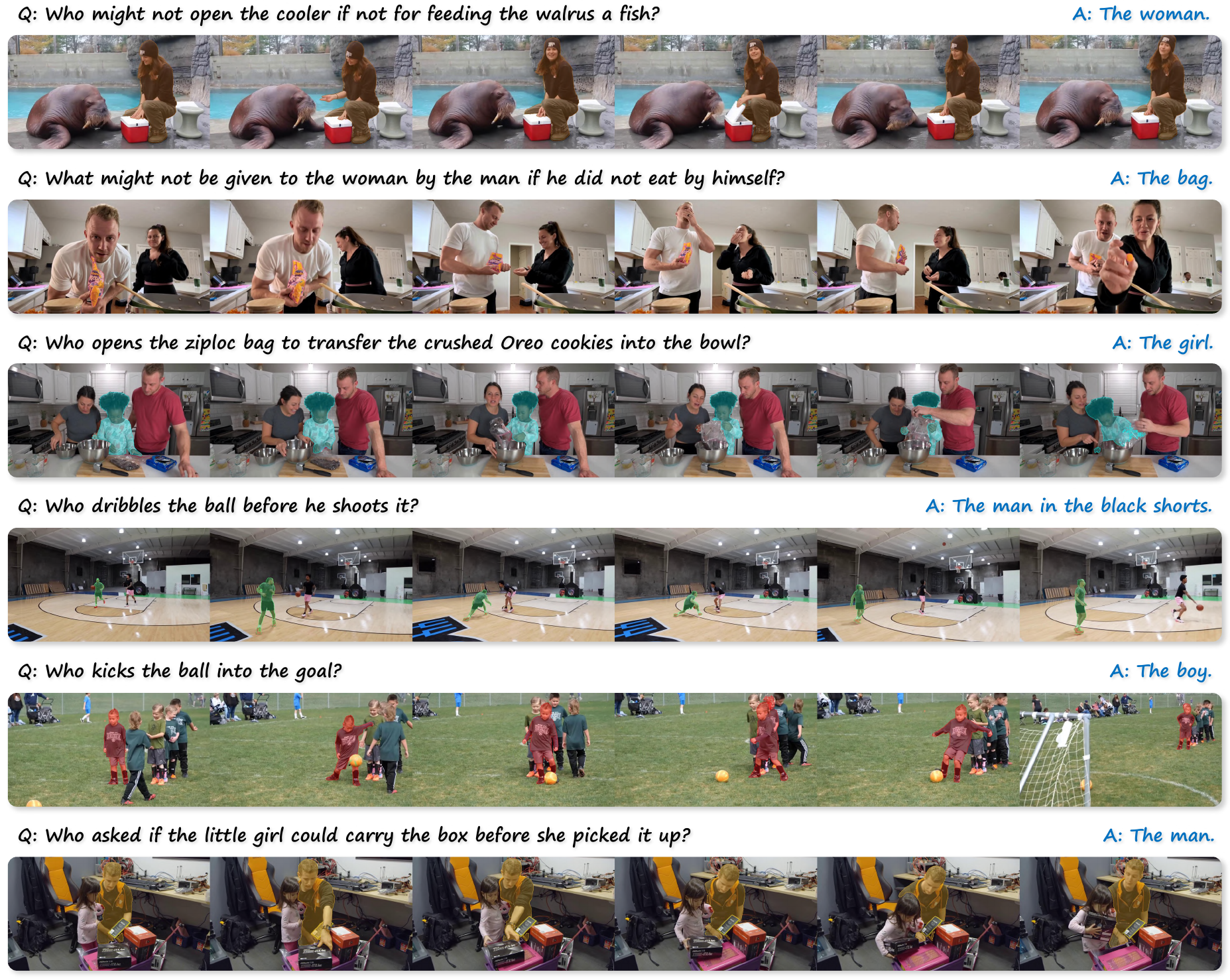}
\caption{Visualization of the predictions from \model{} on GroundMoRe \cite{groundmore}.}
\label{fig:vis_motion}
\end{figure}

\begin{figure}[t]
\centering
\includegraphics[width=\linewidth]{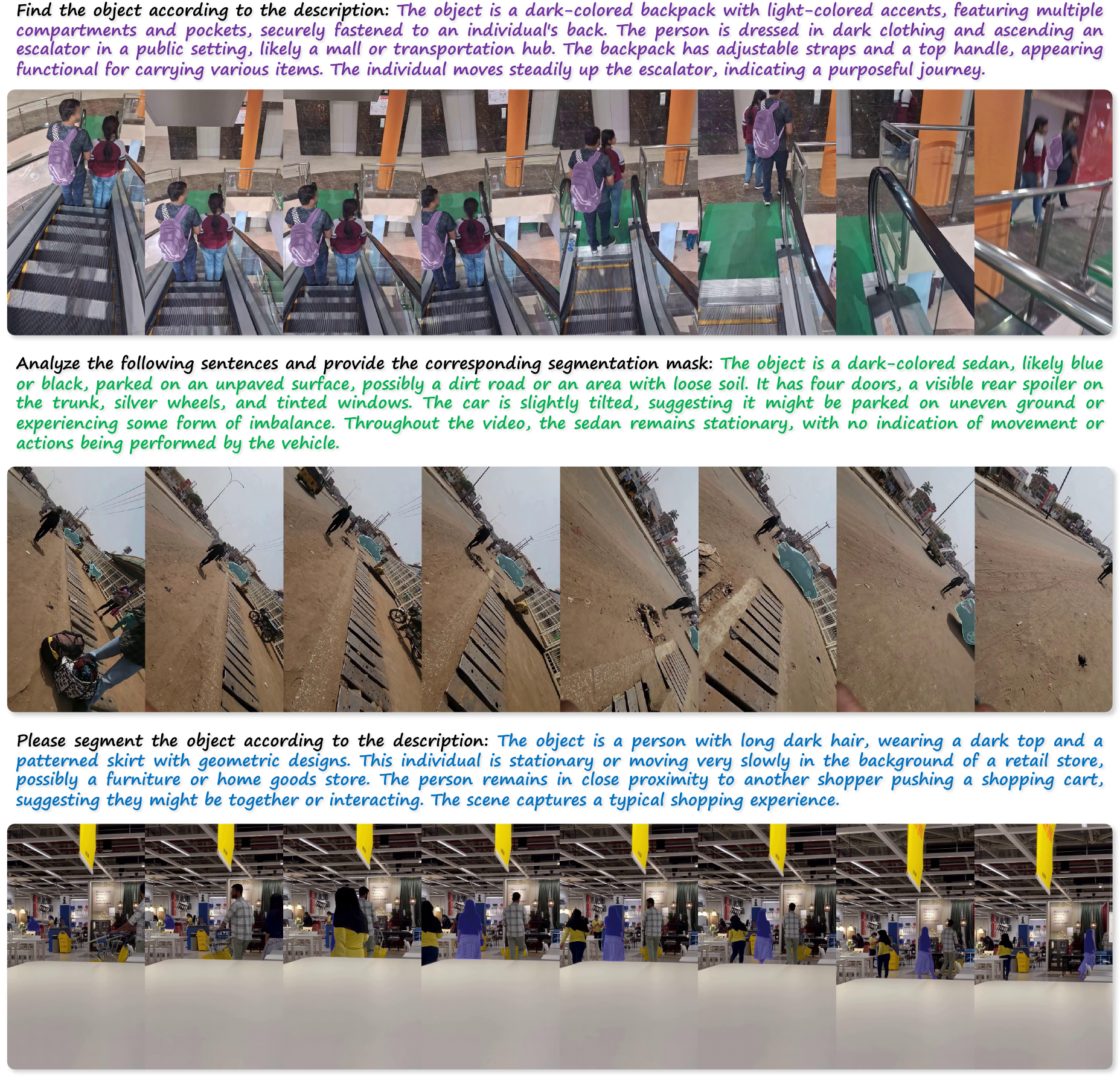}
\caption{Visualization of the predictions from \model{} on Ref-SAV \cite{sa2va}.}
\label{fig:vis_sav}
\end{figure}

\clearpage

\small
\bibliographystyle{plain}
\bibliography{main}

\end{document}